\pdfoutput=1

\documentclass[11pt, british]{article}

\usepackage[final]{acl}

\usepackage{times}
\usepackage{latexsym}
\usepackage{tipa}
\usepackage{tikz}
\usepackage{makecell}
\usepackage{rotating}
\usepackage{amssymb}
\usepackage{pifont} 
\usepackage{tabularray} 
\usepackage{graphicx}
\usepackage{array}
\usepackage{fdsymbol}
\usepackage{cleveref}
\usepackage{booktabs}
\usepackage{amsmath}
\usepackage{multirow}

\newcommand{\cmark}{\ding{51}}%
\newcommand{\xmark}{\ding{55}}%

\usepackage{tcolorbox}
\usepackage{xcolor}
\definecolor{characters}{HTML}{40bcd8}
\definecolor{space}{HTML}{ff6663}
\definecolor{phoneme}{HTML}{ffcf99}

\newtcbox{\characterhighlight}{on line, colback=characters!50, boxrule=0.2mm, left=0.5mm, right=0.2mm, top=0.2mm, bottom=0.2mm}

\newtcbox{\spacehighlight}{on line, colback=space!50, boxrule=0.2mm, left=0.5mm, right=0.2mm, top=0.2mm, bottom=0.2mm}

\newtcbox{\phonemehighlight}{on line, colback=phoneme!50, boxrule=0.2mm, left=0.5mm, right=0.2mm, top=0.2mm, bottom=0.2mm}

\newcommand\mybox[2][]
{\tikz[overlay]\node[fill=blue!10,inner sep=1.3pt, anchor=text, rectangle, rounded corners=1mm,text height=1.5ex, #1] {#2};\phantom{#2}}

\usepackage[T1]{fontenc}

\usepackage[utf8]{inputenc}

\usepackage{microtype}
\usepackage{inconsolata}

\usepackage{twemojis}

\usepackage{tcolorbox}

\usepackage{tabularray}
\usepackage{cellspace}
\newcommand\cincludegraphics[2][]{\raisebox{-0.3\height}{\includegraphics[#1]{#2}}}
\usepackage{setspace}

\title{From Babble to Words \texttwemoji{speech balloon} :\\ Pre-Training Language Models on Continuous Streams of Phonemes}

\author{
    {\bf Z\'{e}bulon Goriely} \texttwemoji{orange} ~~~~~ 
    {\bf Richard Diehl Martinez} \texttwemoji{orange}~~~~ \\
    {\bf Andrew Caines} \texttwemoji{orange}\texttwemoji{lemon} ~~~~
    {\bf Lisa Beinborn} \texttwemoji{green_apple} ~~~~ 
    {\bf Paula Buttery} \texttwemoji{orange}\texttwemoji{lemon} \\
    \texttwemoji{orange} Department of Computer Science \& Technology, University of Cambridge, U.K. \\
    \texttwemoji{lemon} ALTA Institute, University of Cambridge, U.K. \\
    \texttwemoji{green_apple} University of Göttingen, Germany \\ 
    \texttwemoji{orange} \texttt{firstname.secondname@cl.cam.ac.uk} \hspace{2mm}
    \texttwemoji{green_apple} \texttt{lisa.beinborn@uni-goettingen.de}
}

\begin{document}
\maketitle
\begin{abstract}

Language models are typically trained on large corpora of text in their default orthographic form. However, this is not the only option; representing data as streams of phonemes can offer unique advantages, from deeper insights into phonological language acquisition to improved performance on sound-based tasks. The challenge lies in evaluating the impact of phoneme-based training, as most benchmarks are also orthographic. To address this, we develop a pipeline to convert text datasets into a continuous stream of phonemes. We apply this pipeline to the 100-million-word pre-training dataset from the BabyLM challenge, as well as to standard language and grammatical benchmarks, enabling us to pre-train and evaluate a model using phonemic input representations. Our results show that while phoneme-based training slightly reduces performance on traditional language understanding tasks, it offers valuable analytical and practical benefits. 



\begin{tblr}{colspec = {Q[c,m]|X[l,m]}, stretch = 0}
    \cincludegraphics[width=1.4em, keepaspectratio]{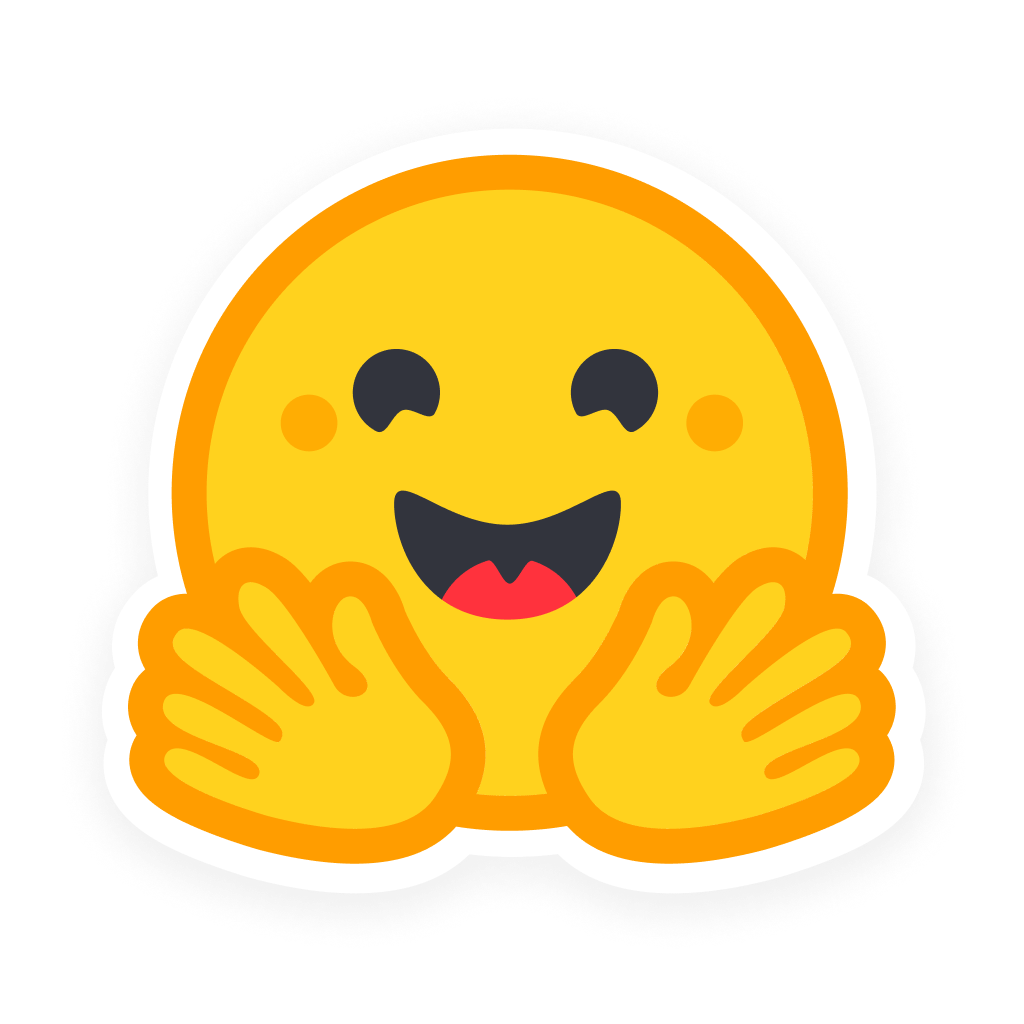} & {\footnotesize{\href{https://huggingface.co/collections/phonemetransformers/from-babble-to-words-66e068b54765a48ff30273c9}{phonemetransformers/FromBabbleToWords}} \tiny{(CC BY 4.0)} } \\
    \cincludegraphics[width=1.2em, keepaspectratio]{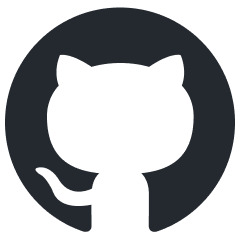} & {\footnotesize{\href{https://github.com/codebyzeb/PhonemeTransformers}{codebyzeb/PhonemeTransformers}} \tiny{(CC BY 4.0)}} \\
    \cincludegraphics[width=1.2em, keepaspectratio]{assets/github-mark.png} & {\footnotesize{\href{https://github.com/codebyzeb/CorpusPhonemizer}{codebyzeb/CorpusPhonemizer}}
    \tiny{(CC BY 4.0)}}
\end{tblr}

\end{abstract}

\vspace{2mm}

\section{Introduction}


The use of orthographic text to train neural networks is so commonplace that it is considered the default. This has not always been the case.

When neural networks were first applied to language, models were primarily trained on continuous streams of phonemes or graphemes, rather than orthographic text with its written artefacts. These early neural models demonstrated a striking ability to acquire phonology, syntax and semantics \citep{elman-1990-finding, seidenberg-1989-word-recognition, prince-1997-optimality}. As technology scaled, subword representations became the dominant representation, offering key advantages such as reducing computation costs and better capturing out-of-vocabulary items \citep{sennrich-etal-2016-bpe}. Written text became favored over speech transcriptions due to matching the domain of downstream tasks and due to the abundance of diverse texts available through web-scraping \citep{bansal-2022-datascaling}. Today, ``large language models'' (LLMs) all use subword-based text inputs and perform impressively on a variety of language understanding tasks \citep{zellers-etal-2019-hellaswag, hendrycks-2020-mmlu, suzgun-2023-Big-Bench}.

\begin{figure}[t]
    \centering
    \includegraphics[width=\linewidth]{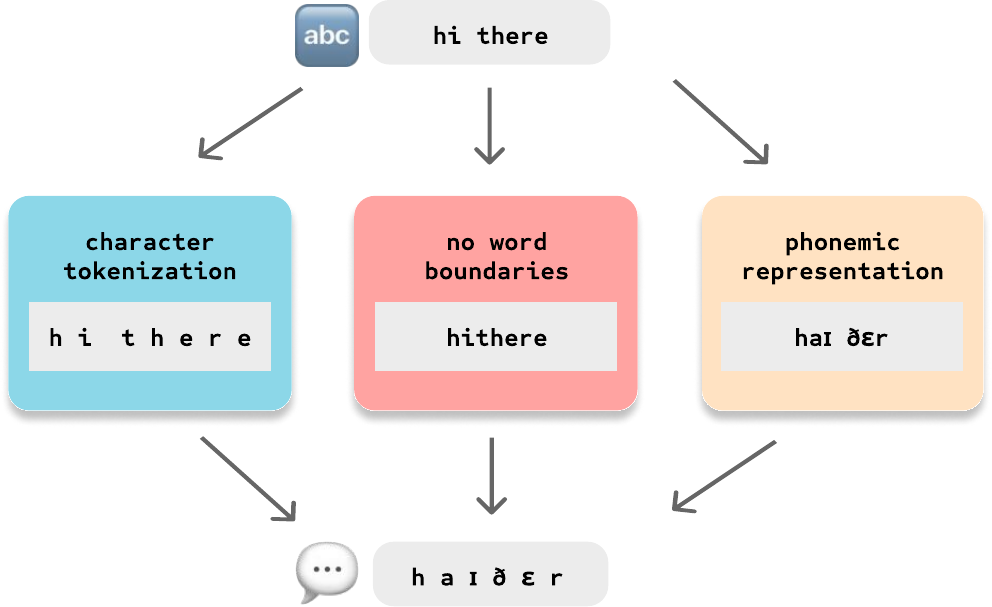}
    \caption{An illustration of all three adjustments that we make to convert text input to continuous streams of phonemes.}
    \label{fig:example}
    \vspace{-6mm}
\end{figure}



The success of these models on downstream tasks has motivated researchers to examine the internal representations of LLMs and analyze their ability to learn grammatical generalizations \citep{hewitt-manning-2019-structural, hu-etal-2020-systematic, manning-2020-emergent}. However, their phonological capabilities remain understudied due to the orthographic nature of training data.



An alternative input representation for text-based language models is to use phonemes rather than graphemes, corresponding to how words are pronounced, rather than how they are written. The use of phonemes, such as those described by the International Phonetic Alphabet (IPA), as an underlying input representation, presents the following analytical and practical benefits over an orthographic representation that is the modern-day default.

\paragraph{Analytical:} A phoneme-based representation is useful when using language models to study the distributional properties of phonemes \citep{mayer-2020-phonology-distribution} and phonological systems of languages more broadly \citep{eden-2018-phonological-distance}. Many language acquisition studies prefer using phonemes as a representation that more closely represents the human learning environment, which facilitates statistical learning experiments ranging from word segmentation \citep{Coltekin2017}, to past-tense formation \citep{kirov-2018-recurrent}, and broader lexico-syntactic knowledge \citep{lavechin}.

\paragraph{Practical:} IPA-encoded text has been found to be beneficial for a variety of NLP tasks including lyric generation \citep{ding-2024-songcomposer}, text-to-speech \citep{sundararaman-2021-phonemebert, li-2023-phoneme-level-bert} and low-resource language modeling \citep{leong-whitenack-2022-phone}. Phonemes also benefit multi-lingual language modeling by establishing a universal representation shared between languages \citep{feng-2023-language-universal-phonetic, zhu-etal-2024-taste}. 




\vspace{.3cm}

Despite the analytical and practical advantages of training language models with phonemes, a key question remains: \emph{Can modern language model architectures encode grammatical knowledge and succeed at language understanding tasks when trained with phoneme-based representations?} 

Answering this question is challenging for two reasons. First, training and evaluation data need to be provided to a model in both a phonemic and graphemic representation. Second, it is non-trivial to select the transformations to convert orthographic text into phonemic representations and to evaluate how these individually affect a model's performance across a wide variety of benchmarks.



In this work, we address these challenges as follows. We first present a method for converting training data and evaluation benchmarks into a unified IPA representation. This enables language models to be trained and evaluated on graphemic and phonemic representations of the same data. We then identify three key transformations which enable us to map from the written representation typically used to train language models to the phonemic representation often used in analytical studies (see \cref{fig:example}). Finally, we conduct a careful ablation of the three transformations: we train a language model on the same corpus of 100 million words with all combinations of the three transformations ($2^3$ configurations), evaluating the model's grammatical capabilities and its resulting performance on downstream language understanding tasks. 


We find that large language models are powerful statistical learners capable of learning grammar from a phonemic input representation. Although we observe a decrease in performance on some tasks, the degradation is not as substantial as has been anecdotally suggested by previous studies. Our ablation studies indicate that the impact of each transformation that we use to convert orthographic text to continuous phoneme streams depends on the downstream task; tasks in the BLiMP Supplement set are particularly sensitive to the use of phonemes, while those in GLUE are sensitive to character tokenization. A deeper analysis into these ablations reveals that many evaluation instances rely on information only present in written text (such as punctuation). Finally, we take advantage of the fact that we train models using phonemic streams and evaluate our models for phonological knowledge using the BabySLM benchmark. Our models achieve the best scores on this benchmark to date.



\section{Related Work}


The standard input representation for training large language models consists of written text split into subword units. By contrast, studies that train models using a phonemic input representation tend to split words into individual phonemes, without word boundaries (as spoken utterances are produced continuously, without clear pauses between words).

We identify three key transformations that bring us from the standard input representation used by language models to this alternative \textbf{phoneme stream} representation:


\begin{itemize}
\setlength\itemsep{0.1em}
    \item \characterhighlight{Character tokenization} Treating each phoneme or grapheme as a token, rather than using subwords.
    \item \spacehighlight{Word boundary removal} Removing whitespace or other word boundary cues from the input.
    \item \phonemehighlight{Phonemic transcription} Converting words to a phonemic representation. 
\end{itemize}




\noindent
Each transformation can be made independently or in combination, as illustrated in \cref{fig:example}. 

Previous literature has extensively explored these three transformations but they have typically been studied independently and been used for different downstream purposes. 

\subsection{Training with Phonemes}

Several language models have been trained with phonemic input \citep{sundararaman-2021-phonemebert, gale-etal-2023-bort} but it remains a challenge to do so due to the lack of large phonemic corpora. While a number of well-known speech-based datasets include phonemic transcriptions, such as Switchboard \citep{godfrey-1992-switchboard} and TIMIT \citep{garofolo-1993-timit}, these datasets are small compared to the trillions of tokens contained in standard language model pre-training corpora \citep{elazar-2024-redpajama}. The majority of works that use phonemic representations typically rely on grapheme to phoneme conversion tools \citep{bisani-2008-g2p, hasegawa-2020-g2pmultilingual} to generate coarse phonemic transliterations of text data.

It is also a challenge to evaluate the broad capabilities of language models trained with phonemes, as most benchmarks assume a graphemic representation, even some that assess phonological knowledge \citep{suvarna-etal-2024-phonologybench}. One benchmark that assesses both the syntactic and phonological capabilities of language models is BabySLM \citep{lavechin}. 
We discuss this benchmark further in \cref{sec:effect}.

\subsection{Character-based Language Models}

The use of characters as the input representation, rather than words or subwords, has been extensively explored. Character-level language models offer a simplified input stream compared to the standard approach of training on learned subword tokens. Many studies have developed specialized architectures to train language models on characters \citep{jozefowicz2016exploringlimitslanguagemodeling, kim2016character, ma-etal-2020-charbert, al-rfou_character-level_2019} while other approaches seek to establish `token-free' training regimes to eliminate the need for subwords entirely \citep{clark-etal-2022-canine, xue-etal-2022-byt5}.

Another alternative input representation is to split words into morphemes, which provide theoretical benefits over subwords and have their own analytical and practical benefits particularly for morphologically rich languages \citep{ustun-etal-2018-characters, nzeyimana-niyongabo-rubungo-2022-kinyabert, fan-sun-2023-constructivist}. Mapping orthographic text to morphemes continues to be a challenging task, requiring dedicated systems trained on labeled corpora \citep{batsuren-etal-2022-sigmorphon} and we do not consider morphemes in this work.

\subsection{Removal of Word Boundaries}

When using a phonemic input representation to model speech, word boundaries are not typically included, as word boundaries are not explicitly marked in the speech stream. The phoneme stream representation (i.e.,\ the combination of all three transformations) is the typical representation for word segmentation studies, where the task is to learn word boundaries without supervision \citep{Brent1999}. A wide variety of statistical, dynamic programming and neural approaches have been applied to the task, with consequences for acquisition research and low-resource language modeling \citep{Blanchard2010, Coltekin2017, algayres_dp-parse_2022, goriely2023word}.

\subsection{Input Representation Comparisons}

To the best of our knowledge, a full systematic comparison of the three input transformations has not yet been conducted.  \citet{hahn-baroni-2019-tabula} investigated the effect of removing word boundaries and using a word-level or character-level tokenization, evaluating on several psycholinguistic benchmarks. However, they only used graphemic text from Wikipedia and did not ablate the two transformations, only comparing a word-level model (with word boundaries) to a character-level model (without word boundaries). \citet{nguyen-2022-word-boundaries} extend this work, comparing character-level graphemic input (with and without word boundaries) to character-level phonemic input (with and without word boundaries) by training on the Librispeech corpus \citep{panayotov2015librispeech}. They also compare larger units of tokenization (BPE and word-level) for both graphemic and phonemic text, but only with word boundaries included, missing out on several key combinations. 

In our work, we provide a complete comparison of these three input representation transformations by considering all combinations, leading to new input representations that have not been studied before (such as subword tokenization trained without word boundaries). We also use a larger model than previous work, a 12-layer transformer rather than a 3-layer LSTM.

\section{Phoneme Stream Pipeline}
\label{sec:pipeline}

To convert the data to a phonemic representation, we developed the \textbf{Corpus Phonemizer} tool:\footnote{\url{https://github.com/codebyzeb/Corpus-Phonemizer}} a library to convert various corpora across many different languages to a unified phonemic representation in IPA, prepare them as Huggingface datasets and subsequently train Huggingface tokenizers.

\subsection{Dataset Phonemization}

Our toolkit leverages the \texttt{phonemizer} package \citep{Bernard2021} with the \texttt{espeak-ng} backend\footnote{\url{https://github.com/espeak-ng/espeak-ng}} which uses a combination of a pronunciation dictionary and pronunciation rules to convert orthographic transcriptions to IPA. We select the American English accent (\texttt{en-US}) for a consistent pronunciation. 

The tool outputs phonemes separated by spaces.\footnote{It is common practice to separate phonemes by spaces to make tokenization simple, as some individual phonemes may consist of several symbols, e.g. \textipa{tS} or \textipa{3I}.} For instance, the phonemic representation of ``what a conundrum!'' is:

\vspace{-2mm}
\begin{center}
\texttt{\textipa{w 2 t} \textvisiblespace~\textipa{2} \textvisiblespace~\textipa{
k @ n 2 n d \*r @ m} \textvisiblespace~}
\end{center}
\vspace{-1mm}

\noindent
One limitation of our phonemization tool is that `a' is not reduced to the shwah, `\textipa{@}', as it would be in continuous speech. We discuss the limitations of this phonemization process in \cref{sec:phonemeslimitations}. Crucially, we lose punctuation marks, as they are an artefact of orthographic text and equivalent information in speech would be conveyed through prosody, stress, or non-linguistic signals such as gestures, none of which are included in this simple phonemic format. This has potential consequences for downstream tasks that rely on such markers, as discussed in \cref{sec:punctuation}.

\begin{table*}[t]
    \centering
    \small
    \addtolength{\tabcolsep}{-0.2em}
    \begin{tabular}{l||ccc|c|c||ccccc}
       Model & \rotatebox[origin=l]{90}{\characterhighlight{Character tokenization}} & \rotatebox[origin=l]{90}{\spacehighlight{Word boundary removal}} & \rotatebox[origin=l]{90}{\phonemehighlight{Phonemic transcription}} & \rotatebox[origin=l]{90}{Vocabulary Size} & Example Tokenization & \rotatebox[origin=l]{90}{BLiMP Filtered} & \rotatebox[origin=l]{90}{BLiMP Supplement} & \rotatebox[origin=l]{90}{GLUE} & \rotatebox[origin=l]{90}{BabySLM (Syntactic)} & \rotatebox[origin=l]{90}{BabySLM ( Lexical)} \\
       \midrule
        Baby Llama & \xmark & \xmark & \xmark & 16,000 & ~\mybox{\textvisiblespace what} ~\mybox{\textvisiblespace a} ~\mybox{\textvisiblespace con} ~\mybox{und} ~\mybox{rum} ~\mybox{\textvisiblespace !} & 73.1 & 60.6 & 69.0 &  94.0 & - \\
        LTG-BERT & \xmark & \xmark & \xmark & 16,000 & ~\mybox{\textvisiblespace what} ~\mybox{\textvisiblespace a} ~\mybox{\textvisiblespace con} ~\mybox{und} ~\mybox{r} ~\mybox{um} ~\mybox{\textvisiblespace !} & 69.3 & 66.5 & 68.4 & 75.8 & - \\
        \midrule
         & \xmark & \xmark & \xmark & 16,000 & ~\mybox{\textvisiblespace what} ~\mybox{\textvisiblespace a} ~\mybox{\textvisiblespace con} ~\mybox{und}  ~\mybox{rum} ~\mybox{\textvisiblespace!} & \textbf{77.8} & \textbf{69.4} & \textbf{71.6} & 92.8 & - \\
         & \xmark & \spacehighlight{\cmark} & \xmark & 16,000 & \mybox{what} ~\mybox{acon} ~\mybox{un} ~\mybox{drum} ~\mybox{!} & 73.9 & 64.3 & 68.6 & 73.9 & - \\
         & \xmark & \xmark & \phonemehighlight{\cmark} & 16,000 & ~\mybox{\textvisiblespace \textipa{w2t}} ~\mybox{\textvisiblespace \textipa{2}} ~\mybox{\textvisiblespace \textipa{k@n}} ~\mybox{\textipa{2nd}} ~\mybox{\textipa{\*r@m}} & 74.7 & 59.6 & 68.6 & 85.8 & 67.3 \\
         \multirow{2}{*}{GPT-2} & \xmark & \spacehighlight{\cmark} & \phonemehighlight{\cmark} & 16,000 & ~\mybox{\textipa{w2t}} ~\mybox{\textipa{2k@n}} ~\mybox{\textipa{2nd}} ~\mybox{\textipa{\*r@m}} & 71.7 & 56.7 & 65.5 & 74.7 & 71.2  \\
         & \characterhighlight{\cmark} & \xmark & \xmark & 115 & \mybox{w} ~\mybox{h} ~\mybox{a} ~\mybox{t} ~\mybox{\textvisiblespace } ~\mybox{a} ~\mybox{\textvisiblespace } ~\mybox{c} ~\mybox{o} ~\mybox{n} ~\mybox{u} ~\mybox{n} ~\mybox{d} ~\mybox{r} ~\mybox{u} ~\mybox{m} ~\mybox{\textvisiblespace } ~\mybox{!} & 77.4 & 63.6 & 64.4 & \textbf{94.9} & - \\
         & \characterhighlight{\cmark} & \spacehighlight{\cmark} & \xmark & 114 & \mybox{w} ~\mybox{h} ~\mybox{a} ~\mybox{t} ~\mybox{a} ~\mybox{c} ~\mybox{o} ~\mybox{n} ~\mybox{u} ~\mybox{n} ~\mybox{d} ~\mybox{r} ~\mybox{u} ~\mybox{m} ~\mybox{!} & 75.1 & 64.8 & 64.8 & 88.3 & - \\
         & \characterhighlight{\cmark} & \xmark & \phonemehighlight{\cmark} & 51 & \mybox{w} ~\mybox{\textipa{2}} ~\mybox{\textipa{t}} ~\mybox{\textvisiblespace } ~\mybox{\textipa{2}}  ~\mybox{\textvisiblespace } ~\mybox{\textipa{k}} ~\mybox{\textipa{@}} ~\mybox{\textipa{n}} ~\mybox{\textipa{2}} ~\mybox{n} ~\mybox{d} ~\mybox{\textipa{\*r}} ~\mybox{\textipa{@}} ~\mybox{m} & 74.7 & 58.5 & 65.6 & 90.5 & \textbf{89.6}  \\
         & \characterhighlight{\cmark} & \spacehighlight{\cmark} & \phonemehighlight{\cmark} & 50 & \mybox{w} ~\mybox{\textipa{2}} ~\mybox{\textipa{t}} ~\mybox{\textipa{2}}  ~\mybox{\textipa{k}} ~\mybox{\textipa{@}} ~\mybox{\textipa{n}} ~\mybox{\textipa{2}} ~\mybox{n} ~\mybox{d} ~\mybox{\textipa{\*r}} ~\mybox{\textipa{@}} ~\mybox{m} & 72.5 & 57.6 & 65.4 & 83.9 & 87.8
    \end{tabular}
    \caption{Results for the two BabyLM baseline models and the GPT-2 model trained under all eight conditions. On the left, we compare the effects of each of the three transformations across all eight possible combinations, by tokenizing the example phrase ``what a conundrum!''. The `\textvisiblespace ' character denotes word boundaries. On the right, we report BLiMP, GLUE and BabySLM scores achieved by each model, with the best scores in each column in \textbf{bold}.} 
    \label{table:results}
    \vspace{-4mm}
\end{table*}

\subsection{Tokenizer Preparation}

Using the phonemic data transcribed by the Corpus Phonemizer tool, our pipeline then implements the three input transformations by preparing different tokenizers:

\begin{itemize}
\setlength\itemsep{0.1em}
    \item \characterhighlight{Character tokenization} We either train the tokenizer using the Byte-Pair Encoding (BPE) algorithm \citep{sennrich-etal-2016-bpe} (\xmark) or create a character-based tokenizer by extracting a vocabulary from the data (\cmark).
    \item \spacehighlight{Word boundary removal} We either train the tokenizer with whitespace included (\xmark) or use the tokenizer's normalizer to strip whitespace (\cmark).  
    \item \phonemehighlight{Phonemic transcription} The tokenizer is either trained on the original orthographic dataset (\xmark), or the phonemized version described above (\cmark).
\end{itemize}




These transformations can be made independently, allowing for all eight combinations of the transformations to be implemented as individual tokenizers. For the combination of BPE and no word boundaries, the whitespace is removed before training, so the model may learn `subwords' that cross word boundaries.

Each tokenizer also adds a dedicated ``utterance boundary'' token \texttt{UTT\_BOUNDARY} to the start of each sentence, representing the pauses between spoken utterances and serving as a dedicated start-of-sentence token. When sentences are collated, it also implicitly acts as an end-of-sentence token, as discussed in \cref{sec:endofsentence}.

\section{Experimental Setup}

We evaluate the effect of our proposed input adjustments by training a GPT-2 model \citep{radford-2019-gpt2} using the BabyLM challenge framework \citep{choshen-et-al-2024-callforpapers-babylm2}. The model is trained eight times with each combination of the three input adjustments. Following the \textsc{strict} track of the BabyLM challenge, we train on a provided corpus of 100 million words and evaluate on a series of benchmarks assessing the grammatical knowledge and the downstream capabilities of each model. We additionally evaluate on BabySLM \citep{lavechin} which provides syntactic and lexical scores specifically for speech-based models. Our phonemized dataset, trained models and tokenizers are hosted on Huggingface.\footnote{\url{https://huggingface.co/collections/phonemetransformers/from-babble-to-words-66e068b54765a48ff30273c9}}

\subsection{Dataset}
\label{sec:input}

The BabyLM 2024 pretraining data contains 100 million words sourced from nine different corpora \citep{warstadt-2023-babylm-findings}. Over 50\% of the data consists of transcribed or scripted speech and over 40\% comes from child-directed sources (written or spoken). 
We apply minor cleaning operations to the dataset, removing extraneous spaces and formatting anomalies using regular expressions.

\subsection{Tokenizers}
\label{sec:tokenizers}

For each of the eight combinations of the three transformations, we train a tokenizer on the `train' portion of the BabyLM dataset.
We compare the output of the eight tokenizers in \cref{table:results}. We used a vocabulary size of 16,000 for the BPE tokenizers to match the vocabulary size used by the two baseline models provided by the BabyLM challenge (described below). 

Note that the vocabulary size for the character-level tokenizers operating on phonemes is less than half the vocabulary size of their orthographic equivalents. This is because the phonemic data only consists of the 47 phonemes produced by the American English accent, but the orthographic data includes numbers, punctuation and other symbols. 

\subsection{Model}
\label{sec:model}

Our experiments use the GPT-2 architecture. We train the model using all eight tokenizers (using the phonemized dataset for the phoneme-based tokenizers) for 400k steps, selecting the checkpoint with the lowest perplexity.\footnote{The best checkpoint for five of the eight models was the final checkpoint but a visual inspection of the curve revealed that differences between the final checkpoints were minimal.} See \cref{app:implementation_details} for a full description of the chosen model parameters and training procedure.

We also report the results from two baseline models which achieved the highest scores at the 2023 BabyLM challenge. These are Baby Llama, an auto-regressive model, which was trained using knowledge distillation from an ensemble of teachers \citep{timiryasov-tastet-2023-baby} and LTG-BERT, an architectural variation of the standard auto-encoding BERT architecture optimized for small, speech-based corpora \citep{samuel-etal-2023-trained, charpentier-samuel-2023-layers}. Both models use a BPE tokenizer with a vocabulary size of 16,000 and have a similar number of parameters to our model.\footnote{Our GPT-2 model has 85M non-embedding parameters. Baby Llama has 41M and LTG-Bert has 110M.}  

\subsection{Evaluation}
\label{sec:evaluation}

We follow the BabyLM Challenge's framework and evaluate on BLiMP \citep{warstadt-etal-2020-blimp-benchmark}, BLiMP Supplement \citep{choshen-et-al-2024-callforpapers-babylm2} and a subset of the (Super)GLUE tasks \citep{wang-etal-2018-glue, wang-etal-2019-superglue}. BLiMP assesses a model's ability to distinguish grammatical sentences from ungrammatical sentences across 67 subtasks covering a range of linguistic phenomena. BLiMP Supplement consists of 5 BLiMP-style tasks covering additional linguistic phenomena not tested by BLiMP. The GLUE suite assesses a language model's language understanding abilities on typical downstream tasks using fine-tuning. 



We also evaluate our models on BabySLM \citep{lavechin}, a benchmark specifically designed for probing speech-based LMs at a \emph{syntactic} level and a \emph{lexical} level. The benchmark was also designed to compare text-based models (those considered here, including both orthographic text and phonemic transcriptions) to speech-based models (which learn directly from audio) by providing parallel text and audio test instances. Finally, the vocabulary items were chosen to be compatible with children’s language experiences, aiming to better reflect the input that children are exposed to as they begin to acquire language. 

The BabySLM syntactic metric is similar to BLiMP, using pairs of grammatical and ungrammatical sentences, but consists of shorter sentences across just six simple syntactic phenomena. By comparison, BLiMP complicated many grammatical phenomena which may be rarely used even in adult--adult spontaneous conversation. 

The lexical metric consists of minimal pairs of words and pseudo-words in a phonemic representation, representing a `real-word recognition' task to assess a model's lexicon and phonemic capabilities. For instance, the model should assign a higher likelihood to the real-word \texttt{\textipa{t~E~m~p~\*r~@~tS~@~\*r}} (temperature) compared to the pseudo-word \texttt{\textipa{t~E~m~p~f~@~tS~@~\*r}} (tempfature). This metric is related to the pronunciation of words, rather than the spelling of words and so cannot be used to evaluate models trained on orthographic text (which have no concept of pronunciation).

To evaluate our phoneme-based models, we run our phonemizer tool on all test instances across these benchmarks (except for the BabySLM lexical examples, which are already in IPA). 

\section{Results}
\label{sec:results}

In \cref{table:results}, we report a summary of the results obtained by the two BabyLM baseline models and our \mbox{GPT-2} model trained in all eight conditions. Due to limited computational resources we only train a single run per condition, limiting our ability to critique them individually. Exact results may be subject to variance across random seeds but we can still observe trends over the whole set. 

The base \mbox{GPT-2} model with no input adjustments outperforms the two baselines for BLiMP, BLiMP Supplement and GLUE, validating our selection of hyper-parameters and choice of architecture as described in \cref{app:implementation_details}.

Comparing the \mbox{GPT-2} model with no input transformations (top row) to the same model with all three transformations applied (bottom row), we notice a decrease in performance across all benchmarks. Although this indicates that the \mbox{GPT-2} architecture is best suited for the standard orthographic input representation (word boundaries, graphemes and subword tokenization), the decrease in performance when the three transformations are applied is not substantial and scores remaining competitive with the baseline models (all combinations still outperform LTG-BERT on BLiMP). It is clear that the model is still capable of learning grammatical rules and excelling at downstream tasks when the input consists of individual phonemes with no word boundaries. 

In \cref{sec:effect} we investigate this result further through an ablation of the three transformations, noting the effect of punctuation and context size. In \cref{sec:babyslm} we focus on the BabySLM metrics, which demonstrate a different pattern to the other benchmarks. Finally, in \cref{sec:punctuation} we investigate the consequences of removing punctuation in our phonemic transcriptions.

\subsection{Teasing Apart the Three Transformations}
\label{sec:effect}

By running our GPT-2 model with all eight combinations of the three input adjustments, we can tease apart the effect of each transformation.

\begin{figure}
    \centering
    \includegraphics[width=0.9\linewidth]{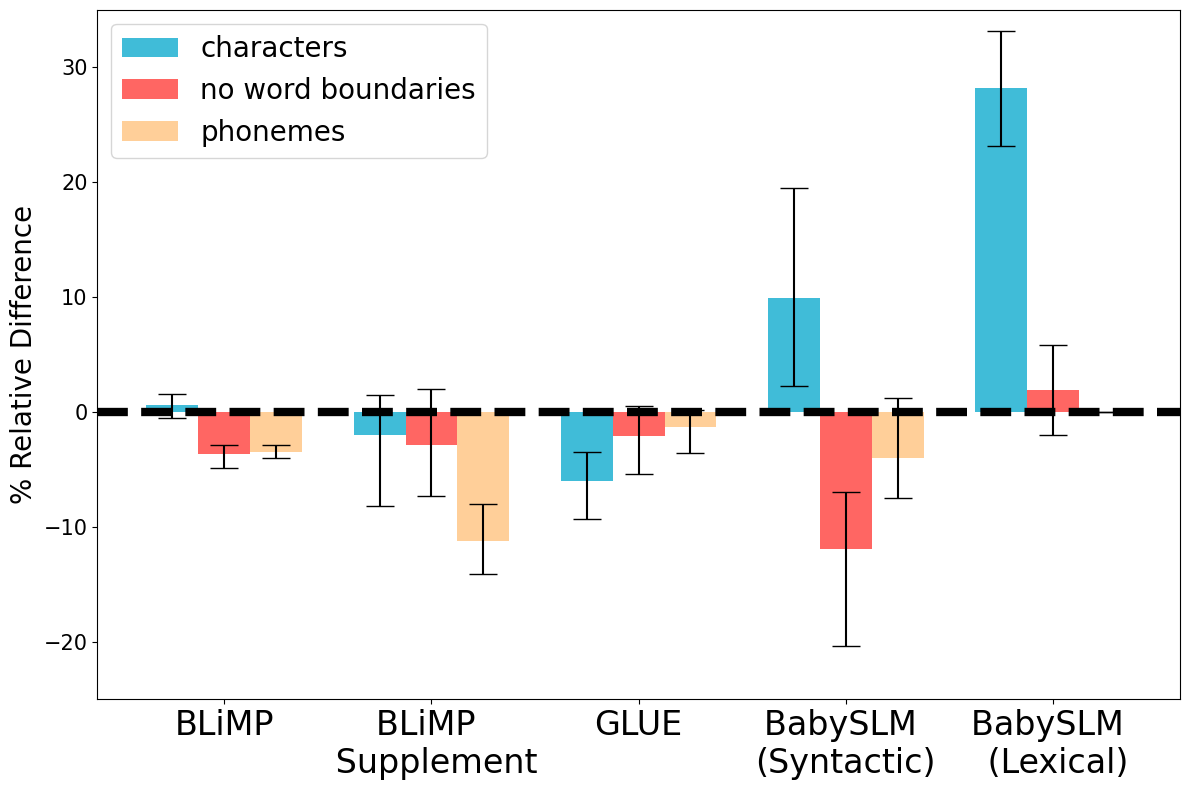}
      \caption{Mean (with Min and Max range) percentage difference achieved on each benchmark's macro score as a result of the three adjustments.}
    \label{fig:condition-differences}
\vspace{-4mm}
\end{figure}

For each transformation, we can create four pairs of runs that only differ with respect to that transformation (e.g.\ the four runs with a phonemic transcription and the four runs with orthographic text). For each pair, we calculate the percentage increase in each metric caused by the transformation. In \cref{fig:condition-differences} we plot the average of these four percentage differences, allowing us to identify the overall effect of each transformation. We can also use the averaged scores for each subtask within a benchmark (such as the 67 BLiMP subtasks) to assess whether differences are significant for BLiMP, BLiMP Supplement, GLUE and BabySLM (Syntactic) using a paired $t$-test (see \cref{sec:significance} for details and $p$-values for each test conducted).

\paragraph{Character Tokenization} We find that character tokenization does not significantly decrease performance on BLiMP or BLiMP Supplement compared to subword tokenization. This validates previous work which found that despite the higher computation costs, character-based language models are just as capable of learning language \citep{al-rfou_character-level_2019, hahn-baroni-2019-tabula}. We do find a significant decrease for GLUE but this may be due to the fact that many of the finetuning examples for GLUE are very long and our model's context size is only 128 tokens, leading to severe truncation. As character-based tokenizers output more tokens for the same sentence than BPE tokenizers, this means that for many GLUE tasks, necessary information is lost.

\paragraph{Word boundary removal} We find that removing word boundaries significantly decreases the BLiMP score, but the decreases for BLiMP Supplement and GLUE are not significant.\footnote{Since there are only 5 tasks for BLiMP Supplement it is difficult to get a $p$-value below 0.05.} In their investigation, \citet{nguyen-2022-word-boundaries} found a decrease of 7-8\% on their own phonemic version of BLiMP when word boundaries were removed, but here we observe only an average decrease of 3.7\%. As they only trained 3-layer LSTMs, it is possible that larger models like ours are required to overcome the loss of word boundaries.

\paragraph{Phonemic Transcription} Finally, we find that using a phonemic transcription instead of the original written text significantly decreases performance on BLiMP and GLUE, although the percentage decreases are small (3.5\% and 1.5\% respectively). It also leads to the largest decrease of 11.3\% for BLiMP Supplement. We discuss a possible explanation for this particular decrease in \cref{sec:punctuation}.

\subsection{BabySLM}
\label{sec:babyslm}

Unlike the other benchmarks, our best BabySLM score is not achieved by the model trained with the standard orthographic input representation. Instead, the best syntactic score of 94.9 is achieved by the model that uses character-based tokenization (on written text, with word boundaries) and the best lexical score of 89.6 is achieved by the model that uses character-based tokenization for phonemes. It is also worth noting that, to the best of our knowledge, these are the best BabySLM scores to date (see \cref{sec:babyslmcomparison} for a detailed comparison).

Examining the effect of each condition, we find that using a phonemic transcription on average reduces the syntactic score by 4.0\%, which is in line with the other benchmarks discussed above. Unlike the other benchmarks, the character tokenization condition \textbf{always leads to an improvement} for both BabySLM scores: an average increase of 9.9\% for the syntactic score and 23.9\% for the lexical score. The sentences used for the syntactic test are all very short compared to the BLiMP sentences (4 words long on average) so a more fine-grained representation may be more useful. For the lexical test, where single words are compared that often only differ by a single phoneme, it seems more appropriate to use a character-based tokenization as the model needs to learn the distributional properties of individual phonemes, which may be lost in subword units. 

The removal of word boundaries has a contrasting effect on the two scores. It reduces the syntactic score by 11.9\% but increases the lexical score by 1.9\%, the only benchmark where removing word boundaries is a positive change. However, the best individual lexical score was achieved by the model that did include word boundaries, suggesting that word boundaries are a helpful signal for a model learning to distinguish words from non-words, possibly because they help separate short sequences of phonemes that appear across word boundaries but not within words. 


For the syntactic score, the worst scores are achieved by the models that learn subwords without word boundaries. For these models, the BPE algorithm is essentially acting as an unsupervised word segmentation algorithm learning to split entire sentences into useful units. With a vocabulary size of 16,000, it seems we learn units smaller than words (morpheme-sized units such as ``un'' in \cref{table:results}) but also units that cross word boundaries (such as ``acon'' in \cref{table:results}). The resulting implicit subword boundaries seem to have particular consequences when evaluating the shorter BabySLM sentences. Using the BPE algorithm in this way could be of interest for word segmentation studies. 

\subsection{The Effect of Punctuation}
\label{sec:punctuation}

Punctuation is a feature of written text that is rarely included in phonemic transcriptions, as it does not typically change the way that words are pronounced. However, punctuation in written text does carry important meaning about the structure and tone of sentences. In speech, this information is typically conveyed through intonation, stress and rhythm. By simply stripping punctuation in our phonemic transcriptions, we may be removing information that is important for a model's ability to learn and process language. 

In some instances, na\"ively stripping punctuation can even lead to nonsense sentences. This may explain the large dip in performance for BLiMP Supplement, as three of the five subtasks rely on punctuation to simulate question-answer pairs or dialogue, such as:

\vspace{-1mm}
\begin{center}
\begin{verbatim}
A: What did you break?\nB: I broke a bowl.
\end{verbatim}
\end{center}
\vspace{-1mm}

In the example above, the line break, colon and question mark are used to indicate speaker turns and convey the question-answer nature of the prompt. Removing the punctuation leads to a nonsense sentence, especially when read aloud with no pauses or change in tone to indicate the structure:

\vspace{-1mm}
\begin{center}
\textipa{2}~\textvisiblespace~\textipa{w 2 t}~\textvisiblespace~\textipa{d I d}~\textvisiblespace~\textipa{j u:}~\textvisiblespace~\textipa{b}~\textipa{\*r}~\textipa{eI k}~\textvisiblespace~\textipa{b i:}~\textvisiblespace~\textipa{aI}~\textvisiblespace\\\textipa{b~\*r~o~U~k}~\textvisiblespace~\textipa{2}~\textvisiblespace~\textipa{b oU l}~\textvisiblespace
\end{center}
\vspace{-1mm}

Without punctuation, the names ``A'' and ``B'' seem out of place. A model trained on written text can use punctuation to possibly understand that these are names, but a spoken model without punctuation would struggle to process this sentence.

This reliance on punctuation seems to be the leading cause of the drop in performance on BLiMP Supplement. If we remove the three subtasks where an understanding of punctuation is required to process the sentence, the effect of switching to a phonemic representation reduces the drop in performance considerably from 11.3\% to 0.9\%.

\begin{figure}[t]
    \centering
    \includegraphics[width=\linewidth]{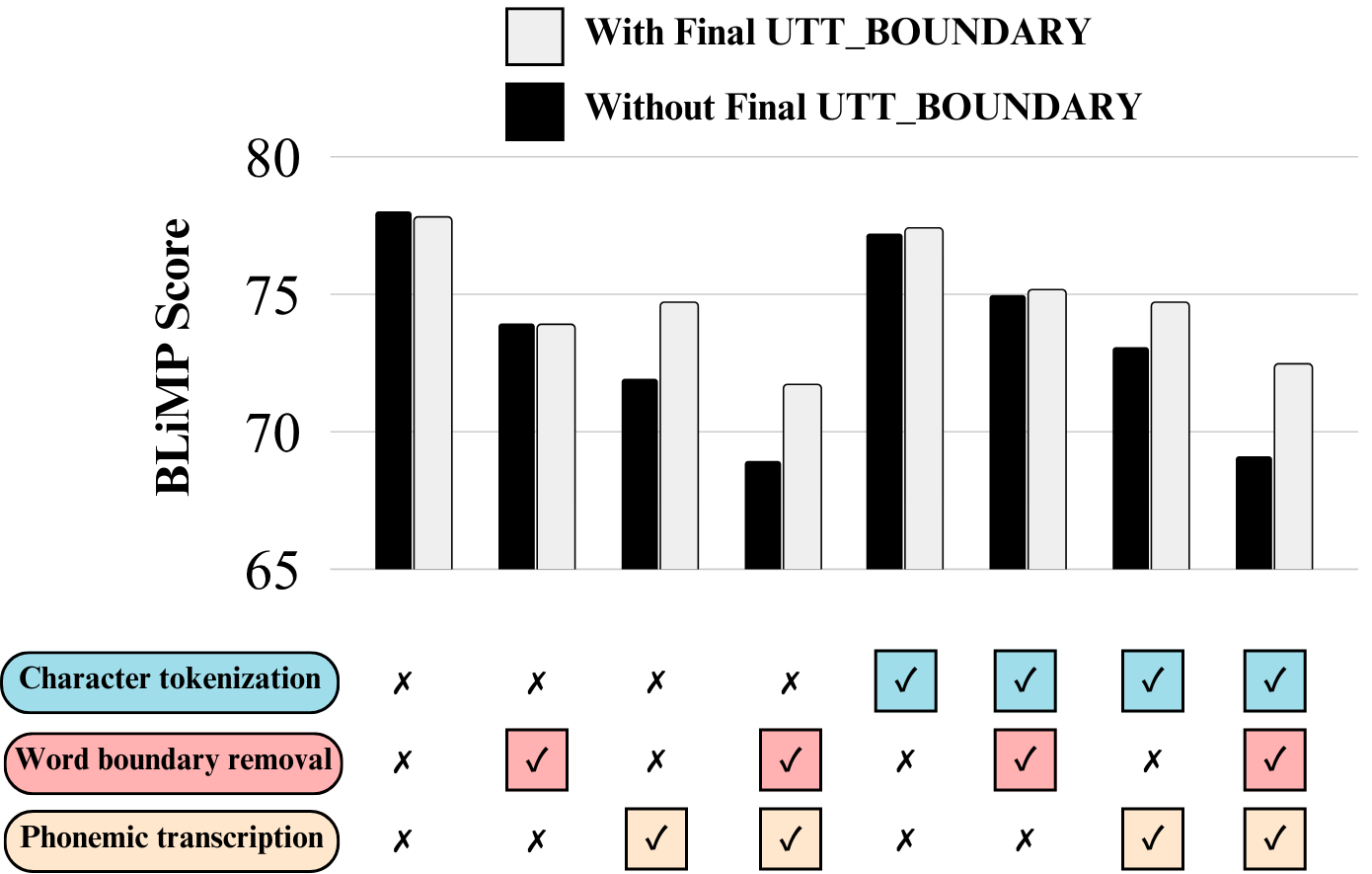}
    \caption{The overall BLiMP scores achieved by GPT-2 in our eight conditions with and without the \texttt{UTT\_BOUNDARY} token (used to separate sentences) included at the end of evaluation instances.}
    \label{fig:endofsentencetoken}
\end{figure}

There is another subtle yet crucial consequence of removing punctuation: stripping punctuation at the end of sentences, if not handled correctly, can lead to significant decreases in performance on these benchmarks. This is because without an end-of-sentence marker, certain evaluation examples are no longer valid. In order to mark the end of the sentences without puncutation, we needed to ensure that our dedicated sentence-separation token was added to the end of each evaluation instance. The effect of this adjustment is highlighted in \cref{fig:endofsentencetoken}. The increase in BLiMP score for our phonemic models confirms that this change was necessary and highlights the importance of carefully investigating the role of tokenization in the evaluation of large language models. We discuss this effect further in \cref{sec:endofsentence}.

\section{Discussion}
\label{sec:gap}

In this work, we set out to establish whether modern language model architectures can encode grammatical knowledge and succeed at language understanding tasks when trained with phonemic input representations. By identifying three key transformations, carefully ablating them and evaluating our models on a wide variety of benchmarks, we found that these transformations do lead to decreased performance on standard benchmarks, but that this decrease is not substantial, and the effect of each transformation varies according to the evaluation. Generally, we conclude that language models are capable learners and training with these input representations is completely viable.

In this section, we consider explanations for the difference in performance across the benchmarks and discuss the limitations of phonemic transcriptions and our monolingual approach. Our work also has implications for human acquisition investigations and studies that train models directly from raw audio, which we discuss in \cref{sec:further}.

\subsection{The Effect of Input Transformations}

There are many possible explanations for the decrease in performance for BLiMP, BLiMP Supplement and GLUE. In \cref{sec:evaluation} and \cref{sec:punctuation} we discuss two possibilities; the fact that character tokenization causes more substantial truncation (affecting GLUE) and the fact that phonemic transcriptions do not include punctuation (which particularly affects BLiMP Supplement). Another factor to consider is that although we do not change the GPT-2 architecture or training parameters, the vocabulary size does change, which affects the size of the embedding layer. Character tokenization also leads to reduced exposure to each sentence during training (fewer epochs) because each sentence is represented with more tokens, increasing the number of steps required for each epoch. Furthermore, our initial choice of model parameters may have implicitly favored the standard orthographic input representation given that the language modeling community has been collectively optimizing these architectures to learn representations for written text, not phonemic streams. Just as the BabyLM challenge seeks to find solutions for low-resource language modeling, we may require an equivalent challenge to identify new methods and architectures for a phonemic input representation. 

We also found a different pattern for the BabySLM benchmark, that certain transformations increased performance. In some cases, the transformations were even necessary (the lexical measure requiring a model to be trained on phonemic input). Given that the BabySLM benchmark more closely relates to child-language acquisition with its shorter sentences and vocabulary taken from child-directed speech, this result will be of interest to studies using language models to study acquisition.

\subsection{Limitations and advantages of phonemic transcriptions}
\label{sec:phonemeslimitations}

One difficulty in training models from ecological long-form child-centered audio is the lack of corpora available. Papers reporting research on day-long recordings tend not to release the raw data due to privacy concerns (e.g.\ \citet{bergelson-etal-2023,leon-cristia-2024}).
Our method allows us to convert text (which is much more readily available) into a speech representation (phoneme streams), meaning that we could quickly prepare a corpus of 100 million words. 

There are also limitations in our transcription generation process. The fact that phonemes are an abstraction of speech means that we lose key information contained in speech such as prosody, stress and allophonic variation. Using a single accent to generate our phonemes, we also lose inter-speaker variability. Children also learn from non-linguistic cues, multi-modal input and interaction. If anything, it is a striking result that a model trained only on a set of 51 discrete symbols is able to demonstrate grammatical knowledge and perform competitively at downstream linguistic tasks. 


\subsection{Multi-lingual evaluation}
\label{sec:limitations} 

A final important remark is that our experiments are conducted only in English. It is possible that language models trained on phonemic data in other languages would exhibit different trends in downstream performance. Although a multilingual analysis is outside the scope of our paper, we have applied our data processing pipeline to prepare phonemized datasets for 31 of the languages contained in the CHILDES database and hope to release this dataset in the near future.

\section{Conclusion}

Our study explores the effect of training language models using phonemic input representations, which offer both analytical and practical advantages. We develop a pipeline to convert orthographic datasets into a continuous stream of phonemes and leverage this pipeline to train a language model on phoneme streams and evaluate its grammatical and language understanding abilities. Our findings suggest that while phoneme-based input representations result in a slight decrease in model performance on traditional language understanding tasks, it is nonetheless a feasible training paradigm, facilitating future language modeling work, improving phonological interpretability and enhancing speech-based applications.  



\section*{Acknowledgements}

Our experiments were performed using resources provided by the Cambridge Service for Data Driven Discovery (CSD3) operated by the University of Cambridge Research Computing Service, provided by Dell EMC and Intel using Tier-2 funding from the Engineering and Physical Sciences Research Council (capital grant EP/T022159/1), and DiRAC funding from the Science and Technology Facilities Council. Z\'ebulon Goriely's work is supported by The Cambridge Trust. Richard Diehl Martinez is supported by the Gates Cambridge Trust (grant OPP1144 from the Bill \& Melinda Gates Foundation). Andrew Caines and Paula Buttery are supported by Cambridge University Press \& Assessment. 
Lisa Beinborn's work is partially supported by the Dutch National Science Organisation (NWO) through the VENI program (Vl.Veni.211C.039).

\bibliographystyle{acl_natbib}

\bibliography{custom}

\begin{thebibliography}{77}
\expandafter\ifx\csname natexlab\endcsname\relax\def\natexlab#1{#1}\fi

\bibitem[{Al-Rfou et~al.(2019)Al-Rfou, Choe, Constant, Guo, and
  Jones}]{al-rfou_character-level_2019}
Rami Al-Rfou, Dokook Choe, Noah Constant, Mandy Guo, and Llion Jones. 2019.
\newblock \href {https://doi.org/10.1609/aaai.v33i01.33013159} {Character-level
  language modeling with deeper self-attention}.
\newblock \emph{Proceedings of the AAAI Conference on Artificial Intelligence},
  33(01):3159--3166.

\bibitem[{Algayres et~al.(2022)Algayres, Ricoul, Karadayi, Laurençon, Zaiem,
  Mohamed, Sagot, and Dupoux}]{algayres_dp-parse_2022}
Robin Algayres, Tristan Ricoul, Julien Karadayi, Hugo Laurençon, Salah Zaiem,
  Abdelrahman Mohamed, Benoît Sagot, and Emmanuel Dupoux. 2022.
\newblock \href {https://doi.org/10.1162/tacl_a_00505} {{DP}-{Parse}: {Finding}
  {Word} {Boundaries} from {Raw} {Speech} with an {Instance} {Lexicon}}.
\newblock \emph{Transactions of the Association for Computational Linguistics},
  10:1051--1065.

\bibitem[{Bansal et~al.(2022)Bansal, Ghorbani, Garg, Zhang, Cherry, Neyshabur,
  and Firat}]{bansal-2022-datascaling}
Yamini Bansal, Behrooz Ghorbani, Ankush Garg, Biao Zhang, Colin Cherry, Behnam
  Neyshabur, and Orhan Firat. 2022.
\newblock Data scaling laws in {NMT}: The effect of noise and architecture.
\newblock In \emph{International Conference on Machine Learning}, pages
  1466--1482. PMLR.

\bibitem[{Baroni(2022)}]{baroni-2022-proper}
Marco Baroni. 2022.
\newblock On the proper role of linguistically oriented deep net analysis in
  linguistic theorising.
\newblock In \emph{Algebraic structures in natural language}, pages 1--16. CRC
  Press.

\bibitem[{Batsuren et~al.(2022)Batsuren, Bella, Arora, Martinovic, Gorman,
  {\v{Z}}abokrtsk{\'y}, Ganbold, Dohnalov{\'a}, {\v{S}}ev{\v{c}}{\'\i}kov{\'a},
  Pelegrinov{\'a}, Giunchiglia, Cotterell, and
  Vylomova}]{batsuren-etal-2022-sigmorphon}
Khuyagbaatar Batsuren, G{\'a}bor Bella, Aryaman Arora, Viktor Martinovic, Kyle
  Gorman, Zden{\v{e}}k {\v{Z}}abokrtsk{\'y}, Amarsanaa Ganbold, {\v{S}}{\'a}rka
  Dohnalov{\'a}, Magda {\v{S}}ev{\v{c}}{\'\i}kov{\'a}, Kate{\v{r}}ina
  Pelegrinov{\'a}, Fausto Giunchiglia, Ryan Cotterell, and Ekaterina Vylomova.
  2022.
\newblock \href {https://doi.org/10.18653/v1/2022.sigmorphon-1.11} {The
  {SIGMORPHON} 2022 shared task on morpheme segmentation}.
\newblock In \emph{Proceedings of the 19th SIGMORPHON Workshop on Computational
  Research in Phonetics, Phonology, and Morphology}, pages 103--116, Seattle,
  Washington. Association for Computational Linguistics.

\bibitem[{Beinborn and Hollenstein(2024)}]{beinborn2024cognitive}
Lisa Beinborn and Nora Hollenstein. 2024.
\newblock \emph{Cognitive plausibility in natural language processing}.
\newblock Springer.

\bibitem[{Bergelson et~al.(2023)Bergelson, Soderstrom, Schwarz, Rowland,
  Ramírez-Esparza, Hamrick, Marklund, Kalashnikova, Guez, Casillas, Benetti,
  van Alphen, and Cristia}]{bergelson-etal-2023}
Elika Bergelson, Melanie Soderstrom, Iris-Corinna Schwarz, Caroline~F. Rowland,
  Nairán Ramírez-Esparza, Lisa~R. Hamrick, Ellen Marklund, Marina
  Kalashnikova, Ava Guez, Marisa Casillas, Lucia Benetti, Petra van Alphen, and
  Alejandrina Cristia. 2023.
\newblock \href {https://doi.org/10.1073/pnas.2300671120} {Everyday language
  input and production in 1,001 children from six continents}.
\newblock \emph{Proceedings of the National Academy of Sciences},
  120(52):e2300671120.

\bibitem[{Bernard and Titeux(2021)}]{Bernard2021}
Mathieu Bernard and Hadrien Titeux. 2021.
\newblock \href {https://doi.org/10.21105/joss.03958} {Phonemizer: Text to
  phones transcription for multiple languages in python}.
\newblock \emph{Journal of Open Source Software}, 6(68):3958.

\bibitem[{Biderman et~al.(2023)Biderman, Schoelkopf, Anthony, Bradley,
  O’Brien, Hallahan, Khan, Purohit, Prashanth, Raff
  et~al.}]{biderman2023pythia}
Stella Biderman, Hailey Schoelkopf, Quentin~Gregory Anthony, Herbie Bradley,
  Kyle O’Brien, Eric Hallahan, Mohammad~Aflah Khan, Shivanshu Purohit,
  USVSN~Sai Prashanth, Edward Raff, et~al. 2023.
\newblock Pythia: A suite for analyzing large language models across training
  and scaling.
\newblock In \emph{International Conference on Machine Learning}, pages
  2397--2430.

\bibitem[{Bisani and Ney(2008)}]{bisani-2008-g2p}
Maximilian Bisani and Hermann Ney. 2008.
\newblock Joint-sequence models for grapheme-to-phoneme conversion.
\newblock \emph{Speech communication}, 50(5):434--451.

\bibitem[{Blanchard et~al.(2010)Blanchard, Heinz, and
  Golinkoff}]{Blanchard2010}
Daniel Blanchard, Jeffrey Heinz, and Roberta Golinkoff. 2010.
\newblock \href {https://doi.org/10.1017/S030500090999050X} {{Modeling the
  contribution of phonotactic cues to the problem of word segmentation}}.
\newblock \emph{Journal of Child Language}, 37(3):487--511.

\bibitem[{B{\"o}rschinger et~al.(2013)B{\"o}rschinger, Johnson, and
  Demuth}]{borschinger-etal-2013-joint}
Benjamin B{\"o}rschinger, Mark Johnson, and Katherine Demuth. 2013.
\newblock \href {https://aclanthology.org/P13-1148} {A joint model of word
  segmentation and phonological variation for {E}nglish word-final
  /t/-deletion}.
\newblock In \emph{Proceedings of the 51st Annual Meeting of the Association
  for Computational Linguistics (Volume 1: Long Papers)}, pages 1508--1516,
  Sofia, Bulgaria. Association for Computational Linguistics.

\bibitem[{Brent(1999)}]{Brent1999}
Michael~R. Brent. 1999.
\newblock \href {https://doi.org/10.1023/a:1007541817488} {{Efficient,
  probabilistically sound algorithm for segmentation and word discovery}}.
\newblock \emph{Machine Learning}, 34(1):71--105.

\bibitem[{Charpentier and Samuel(2023)}]{charpentier-samuel-2023-layers}
Lucas Georges~Gabriel Charpentier and David Samuel. 2023.
\newblock \href {https://doi.org/10.18653/v1/2023.conll-babylm.20} {Not all
  layers are equally as important: Every layer counts {BERT}}.
\newblock In \emph{Proceedings of the BabyLM Challenge at the 27th Conference
  on Computational Natural Language Learning}, pages 238--252, Singapore.
  Association for Computational Linguistics.

\bibitem[{Choshen et~al.(2024)Choshen, Cotterell, Hu, Linzen, Mueller, Ross,
  Warstadt, Wilcox, Williams, and
  Zhuang}]{choshen-et-al-2024-callforpapers-babylm2}
Leshem Choshen, Ryan Cotterell, Michael~Y Hu, Tal Linzen, Aaron Mueller,
  Candace Ross, Alex Warstadt, Ethan Wilcox, Adina Williams, and Chengxu
  Zhuang. 2024.
\newblock Call for papers -- {The BabyLM Challenge}: Sample-efficient
  pretraining on a developmentally plausible corpus.
\newblock \emph{arXiv preprint arXiv:2404.06214}.

\bibitem[{Clark et~al.(2022)Clark, Garrette, Turc, and
  Wieting}]{clark-etal-2022-canine}
Jonathan~H. Clark, Dan Garrette, Iulia Turc, and John Wieting. 2022.
\newblock \href {https://doi.org/10.1162/tacl_a_00448} {Canine: Pre-training an
  efficient tokenization-free encoder for language representation}.
\newblock \emph{Transactions of the Association for Computational Linguistics},
  10:73--91.

\bibitem[{{\c{C}}{\"{o}}ltekin(2017)}]{Coltekin2017}
{\c{C}}ağrı {\c{C}}{\"{o}}ltekin. 2017.
\newblock \href {https://doi.org/10.1111/cogs.12454} {{Using Predictability for
  Lexical Segmentation}}.
\newblock \emph{Cognitive Science}, 41(7):1988--2021.

\bibitem[{Ding et~al.(2024)Ding, Liu, Dong, Zhang, Qian, He, Lin, and
  Wang}]{ding-2024-songcomposer}
Shuangrui Ding, Zihan Liu, Xiaoyi Dong, Pan Zhang, Rui Qian, Conghui He, Dahua
  Lin, and Jiaqi Wang. 2024.
\newblock Songcomposer: A large language model for lyric and melody composition
  in song generation.
\newblock \emph{arXiv preprint arXiv:2402.17645}.

\bibitem[{Dunbar et~al.(2022)Dunbar, Hamilakis, and
  Dupoux}]{dunbar_self-supervised_2022}
Ewan Dunbar, Nicolas Hamilakis, and Emmanuel Dupoux. 2022.
\newblock \href {https://doi.org/10.1109/JSTSP.2022.3206084} {Self-{Supervised}
  {Language} {Learning} {From} {Raw} {Audio}: {Lessons} {From} the {Zero}
  {Resource} {Speech} {Challenge}}.
\newblock \emph{IEEE Journal of Selected Topics in Signal Processing},
  16(6):1211--1226.
\newblock Conference Name: IEEE Journal of Selected Topics in Signal
  Processing.

\bibitem[{Dupoux(2018)}]{dupoux-2018-cognitive}
Emmanuel Dupoux. 2018.
\newblock Cognitive science in the era of artificial intelligence: A roadmap
  for reverse-engineering the infant language-learner.
\newblock \emph{Cognition}, 173:43--59.

\bibitem[{Eden(2018)}]{eden-2018-phonological-distance}
S~Elizabeth Eden. 2018.
\newblock \emph{Measuring phonological distance between languages}.
\newblock Ph.D. thesis, UCL (University College London).

\bibitem[{Elazar et~al.(2024)Elazar, Bhagia, Magnusson, Ravichander, Schwenk,
  Suhr, Walsh, Groeneveld, Soldaini, Singh et~al.}]{elazar-2024-redpajama}
Yanai Elazar, Akshita Bhagia, Ian~Helgi Magnusson, Abhilasha Ravichander,
  Dustin Schwenk, Alane Suhr, Evan~Pete Walsh, Dirk Groeneveld, Luca Soldaini,
  Sameer Singh, et~al. 2024.
\newblock What's in my big data?
\newblock In \emph{The Twelfth International Conference on Learning
  Representations}.

\bibitem[{Elman(1990)}]{elman-1990-finding}
Jeffrey~L Elman. 1990.
\newblock Finding structure in time.
\newblock \emph{Cognitive science}, 14(2):179--211.

\bibitem[{Evanson et~al.(2023)Evanson, Lakretz, and
  King}]{evanson-etal-2023-language}
Linnea Evanson, Yair Lakretz, and Jean~R{\'e}mi King. 2023.
\newblock \href {https://doi.org/10.18653/v1/2023.findings-acl.773} {Language
  acquisition: do children and language models follow similar learning stages?}
\newblock In \emph{Findings of the Association For Computational Linguistics:
  ACL 2023}, pages 12205--12218, Toronto, Canada. Association for Computational
  Linguistics.

\bibitem[{Fan and Sun(2023)}]{fan-sun-2023-constructivist}
Allison Fan and Weiwei Sun. 2023.
\newblock \href {https://aclanthology.org/2023.cxgsnlp-1.5} {Constructivist
  tokenization for {E}nglish}.
\newblock In \emph{Proceedings of the First International Workshop on
  Construction Grammars and NLP (CxGs+NLP, GURT/SyntaxFest 2023)}, pages
  36--40, Washington, D.C. Association for Computational Linguistics.

\bibitem[{Feliciano~de
  Faria(2019)}]{feliciano-de-faria-2019-utterance-boundaries}
Pablo~Picasso Feliciano~de Faria. 2019.
\newblock \href {https://doi.org/10.18653/v1/W19-2917} {The role of utterance
  boundaries and word frequencies for part-of-speech learning in {B}razilian
  {P}ortuguese through distributional analysis}.
\newblock In \emph{Proceedings of the Workshop on Cognitive Modeling and
  Computational Linguistics}, pages 152--159, Minneapolis, Minnesota.
  Association for Computational Linguistics.

\bibitem[{Feng et~al.(2023)Feng, Tu, Xia, Huang, and
  Wang}]{feng-2023-language-universal-phonetic}
Siyuan Feng, Ming Tu, Rui Xia, Chuanzeng Huang, and Yuxuan Wang. 2023.
\newblock Language-universal phonetic representation in multilingual speech
  pretraining for low-resource speech recognition.
\newblock In \emph{{INTERSPEECH 2023}}, Dublin, Ireland. {ISCA}.

\bibitem[{Gale et~al.(2023)Gale, Salem, Fergadiotis, and
  Bedrick}]{gale-etal-2023-bort}
Robert Gale, Alexandra Salem, Gerasimos Fergadiotis, and Steven Bedrick. 2023.
\newblock \href {https://doi.org/10.18653/v1/2023.repl4nlp-1.18} {Mixed
  orthographic/phonemic language modeling: Beyond orthographically restricted
  transformers ({BORT})}.
\newblock In \emph{Proceedings of the 8th Workshop on Representation Learning
  for NLP (RepL4NLP 2023)}, pages 212--225, Toronto, Canada. Association for
  Computational Linguistics.

\bibitem[{Garofolo et~al.(1993)Garofolo, Lamel, Fisher, Fiscus, and
  Pallett}]{garofolo-1993-timit}
John~S Garofolo, Lori~F Lamel, William~M Fisher, Jonathan~G Fiscus, and David~S
  Pallett. 1993.
\newblock Darpa timit acoustic-phonetic continous speech corpus cd-rom. nist
  speech disc 1-1.1.
\newblock \emph{NASA STI/Recon technical report n}, 93:27403.

\bibitem[{Godfrey et~al.(1992)Godfrey, Holliman, and
  McDaniel}]{godfrey-1992-switchboard}
John~J Godfrey, Edward~C Holliman, and Jane McDaniel. 1992.
\newblock Switchboard: Telephone speech corpus for research and development.
\newblock In \emph{Acoustics, speech, and signal processing, ieee international
  conference on}, volume~1, pages 517--520. IEEE Computer Society.

\bibitem[{Goriely et~al.(2023)Goriely, Caines, and Buttery}]{goriely2023word}
Z{\'e}bulon Goriely, Andrew Caines, and Paula Buttery. 2023.
\newblock Word segmentation from transcriptions of child-directed speech using
  lexical and sub-lexical cues.
\newblock \emph{Journal of Child Language}, pages 1--41.

\bibitem[{Hahn and Baroni(2019)}]{hahn-baroni-2019-tabula}
Michael Hahn and Marco Baroni. 2019.
\newblock \href {https://doi.org/10.1162/tacl_a_00283} {Tabula nearly rasa:
  Probing the linguistic knowledge of character-level neural language models
  trained on unsegmented text}.
\newblock \emph{Transactions of the Association for Computational Linguistics},
  7:467--484.

\bibitem[{Hasegawa-Johnson et~al.(2020)Hasegawa-Johnson, Rolston, Goudeseune,
  Levow, and Kirchhoff}]{hasegawa-2020-g2pmultilingual}
Mark Hasegawa-Johnson, Leanne Rolston, Camille Goudeseune, Gina-Anne Levow, and
  Katrin Kirchhoff. 2020.
\newblock Grapheme-to-phoneme transduction for cross-language asr.
\newblock In \emph{International Conference on Statistical Language and Speech
  Processing}, pages 3--19. Springer.

\bibitem[{Hendrycks et~al.(2020)Hendrycks, Burns, Basart, Zou, Mazeika, Song,
  and Steinhardt}]{hendrycks-2020-mmlu}
Dan Hendrycks, Collin Burns, Steven Basart, Andy Zou, Mantas Mazeika, Dawn
  Song, and Jacob Steinhardt. 2020.
\newblock Measuring massive multitask language understanding.
\newblock In \emph{International Conference on Learning Representations}.

\bibitem[{Hewitt and Manning(2019)}]{hewitt-manning-2019-structural}
John Hewitt and Christopher~D. Manning. 2019.
\newblock \href {https://doi.org/10.18653/v1/N19-1419} {{A} structural probe
  for finding syntax in word representations}.
\newblock In \emph{Proceedings of the 2019 Conference of the North {A}merican
  Chapter of the Association for Computational Linguistics: Human Language
  Technologies, Volume 1 (Long and Short Papers)}, pages 4129--4138,
  Minneapolis, Minnesota. Association for Computational Linguistics.

\bibitem[{Hollenstein et~al.(2021)Hollenstein, Pirovano, Zhang, J{\"a}ger, and
  Beinborn}]{hollenstein-etal-2021-multilingual}
Nora Hollenstein, Federico Pirovano, Ce~Zhang, Lena J{\"a}ger, and Lisa
  Beinborn. 2021.
\newblock \href {https://doi.org/10.18653/v1/2021.naacl-main.10} {Multilingual
  language models predict human reading behavior}.
\newblock In \emph{Proceedings of the 2021 Conference of the North American
  Chapter of the Association for Computational Linguistics: Human Language
  Technologies,}, pages 106--123, Online. Association for Computational
  Linguistics.

\bibitem[{Hu et~al.(2020)Hu, Gauthier, Qian, Wilcox, and
  Levy}]{hu-etal-2020-systematic}
Jennifer Hu, Jon Gauthier, Peng Qian, Ethan Wilcox, and Roger Levy. 2020.
\newblock \href {https://doi.org/10.18653/v1/2020.acl-main.158} {A systematic
  assessment of syntactic generalization in neural language models}.
\newblock In \emph{Proceedings of the 58th Annual Meeting of the Association
  for Computational Linguistics}, pages 1725--1744, Online. Association for
  Computational Linguistics.

\bibitem[{Huebner et~al.(2021)Huebner, Sulem, Cynthia, and
  Roth}]{huebner-etal-2021-babyberta}
Philip~A. Huebner, Elior Sulem, Fisher Cynthia, and Dan Roth. 2021.
\newblock \href {https://doi.org/10.18653/v1/2021.conll-1.49} {{B}aby{BERT}a:
  Learning more grammar with small-scale child-directed language}.
\newblock In \emph{Proceedings of the 25th Conference on Computational Natural
  Language Learning}, pages 624--646, Online. Association for Computational
  Linguistics.

\bibitem[{Jozefowicz et~al.(2016)Jozefowicz, Vinyals, Schuster, Shazeer, and
  Wu}]{jozefowicz2016exploringlimitslanguagemodeling}
Rafal Jozefowicz, Oriol Vinyals, Mike Schuster, Noam Shazeer, and Yonghui Wu.
  2016.
\newblock \href {http://arxiv.org/abs/1602.02410} {Exploring the limits of
  language modeling}.

\bibitem[{Kahn et~al.(2020)Kahn, Riviere, Zheng, Kharitonov, Xu, Mazar{\'e},
  Karadayi, Liptchinsky, Collobert, Fuegen et~al.}]{kahn2020libri}
Jacob Kahn, Morgane Riviere, Weiyi Zheng, Evgeny Kharitonov, Qiantong Xu,
  Pierre-Emmanuel Mazar{\'e}, Julien Karadayi, Vitaliy Liptchinsky, Ronan
  Collobert, Christian Fuegen, et~al. 2020.
\newblock Libri-light: A benchmark for asr with limited or no supervision.
\newblock In \emph{ICASSP 2020-2020 IEEE International Conference on Acoustics,
  Speech and Signal Processing (ICASSP)}, pages 7669--7673.

\bibitem[{Kazanina et~al.(2018)Kazanina, Bowers, and
  Idsardi}]{kazanina2018phonemes}
Nina Kazanina, Jeffrey~S Bowers, and William Idsardi. 2018.
\newblock Phonemes: Lexical access and beyond.
\newblock \emph{Psychonomic bulletin \& review}, 25(2):560--585.

\bibitem[{Kim et~al.(2016)Kim, Jernite, Sontag, and Rush}]{kim2016character}
Yoon Kim, Yacine Jernite, David Sontag, and Alexander Rush. 2016.
\newblock Character-aware neural language models.
\newblock In \emph{Proceedings of the AAAI conference on artificial
  intelligence}, volume~30.

\bibitem[{Kirov and Cotterell(2018)}]{kirov-2018-recurrent}
Christo Kirov and Ryan Cotterell. 2018.
\newblock \href {https://doi.org/10.1162/tacl_a_00247} {{Recurrent Neural
  Networks in Linguistic Theory: Revisiting Pinker and Prince (1988) and the
  Past Tense Debate}}.
\newblock \emph{Transactions of the Association for Computational Linguistics},
  6:651--665.

\bibitem[{Lavechin et~al.(2022)Lavechin, De~Seyssel, Titeux, Bredin,
  Wisniewski, Cristia, and Dupoux}]{lavechin2022can}
Marvin Lavechin, Maureen De~Seyssel, Hadrien Titeux, Herv{\'e} Bredin,
  Guillaume Wisniewski, Alejandrina Cristia, and Emmanuel Dupoux. 2022.
\newblock Can statistical learning bootstrap early language acquisition? a
  modeling investigation.

\bibitem[{Lavechin et~al.(2023)Lavechin, Sy, Titeux, Bland{\'o}n,
  R{\"a}s{\"a}nen, Bredin, Dupoux, and Cristia}]{lavechin}
Marvin Lavechin, Yaya Sy, Hadrien Titeux, Mar{\'i}a Andrea~Cruz Bland{\'o}n,
  Okko R{\"a}s{\"a}nen, Herv{\'e} Bredin, Emmanuel Dupoux, and Alejandrina
  Cristia. 2023.
\newblock \href {https://doi.org/10.21437/Interspeech.2023-978} {{BabySLM:
  language-acquisition-friendly benchmark of self-supervised spoken language
  models}}.
\newblock In \emph{{INTERSPEECH 2023}}, pages 4588--4592, Dublin, Ireland.
  {ISCA}.

\bibitem[{Leong and Whitenack(2022)}]{leong-whitenack-2022-phone}
Colin Leong and Daniel Whitenack. 2022.
\newblock \href {https://doi.org/10.18653/v1/2022.acl-long.364} {Phone-ing it
  in: Towards flexible multi-modal language model training by phonetic
  representations of data}.
\newblock In \emph{Proceedings of the 60th Annual Meeting of the Association
  for Computational Linguistics (Volume 1: Long Papers)}, pages 5306--5315,
  Dublin, Ireland. Association for Computational Linguistics.

\bibitem[{Li et~al.(2023)Li, Han, Jiang, and
  Mesgarani}]{li-2023-phoneme-level-bert}
Yinghao~Aaron Li, Cong Han, Xilin Jiang, and Nima Mesgarani. 2023.
\newblock Phoneme-level {BERT} for enhanced prosody of text-to-speech with
  grapheme predictions.
\newblock In \emph{ICASSP 2023-2023 IEEE International Conference on Acoustics,
  Speech and Signal Processing (ICASSP)}, pages 1--5.

\bibitem[{Léon and Cristia(2024)}]{leon-cristia-2024}
Mathilde Léon and Alejandrina Cristia. 2024.
\newblock \href {https://doi.org/10.31219/osf.io/dy4wt} {\emph{Data Protection
  Handbook for Long-Form Recording Research: Navigating Data Protection Laws
  across the Globe}}.
\newblock OSF.

\bibitem[{Ma et~al.(2020)Ma, Cui, Si, Liu, Wang, and
  Hu}]{ma-etal-2020-charbert}
Wentao Ma, Yiming Cui, Chenglei Si, Ting Liu, Shijin Wang, and Guoping Hu.
  2020.
\newblock \href {https://doi.org/10.18653/v1/2020.coling-main.4} {{C}har{BERT}:
  Character-aware pre-trained language model}.
\newblock In \emph{Proceedings of the 28th International Conference on
  Computational Linguistics}, pages 39--50, Barcelona, Spain (Online).
  International Committee on Computational Linguistics.

\bibitem[{MacWhinney and Snow(1985)}]{macwhinney1985child}
Brian MacWhinney and Catherine Snow. 1985.
\newblock {The Child Language Data Exchange System}.
\newblock \emph{Journal of Child Language}, 12(2):271--295.

\bibitem[{Manning et~al.(2020)Manning, Clark, Hewitt, Khandelwal, and
  Levy}]{manning-2020-emergent}
Christopher~D Manning, Kevin Clark, John Hewitt, Urvashi Khandelwal, and Omer
  Levy. 2020.
\newblock Emergent linguistic structure in artificial neural networks trained
  by self-supervision.
\newblock \emph{Proceedings of the National Academy of Sciences},
  117(48):30046--30054.

\bibitem[{Matusevych et~al.(2023)Matusevych, Schatz, Kamper, Feldman, and
  Goldwater}]{matusevych2023infant}
Yevgen Matusevych, Thomas Schatz, Herman Kamper, Naomi~H Feldman, and Sharon
  Goldwater. 2023.
\newblock Infant phonetic learning as perceptual space learning: A
  crosslinguistic evaluation of computational models.
\newblock \emph{Cognitive Science}, 47(7):e13314.

\bibitem[{Mayer(2020)}]{mayer-2020-phonology-distribution}
Connor Mayer. 2020.
\newblock An algorithm for learning phonological classes from distributional
  similarity.
\newblock \emph{Phonology}, 37(1):91--131.

\bibitem[{Nguyen et~al.(2022)Nguyen, De~Seyssel, Algayres, Roze, Dunbar, and
  Dupoux}]{nguyen-2022-word-boundaries}
Tu~Anh Nguyen, Maureen De~Seyssel, Robin Algayres, Patricia Roze, Ewan Dunbar,
  and Emmanuel Dupoux. 2022.
\newblock Are word boundaries useful for unsupervised language learning?
\newblock \emph{arXiv preprint arXiv:2210.02956}.

\bibitem[{Nzeyimana and
  Niyongabo~Rubungo(2022)}]{nzeyimana-niyongabo-rubungo-2022-kinyabert}
Antoine Nzeyimana and Andre Niyongabo~Rubungo. 2022.
\newblock \href {https://doi.org/10.18653/v1/2022.acl-long.367} {{K}inya{BERT}:
  a morphology-aware {K}inyarwanda language model}.
\newblock In \emph{Proceedings of the 60th Annual Meeting of the Association
  for Computational Linguistics (Volume 1: Long Papers)}, pages 5347--5363,
  Dublin, Ireland. Association for Computational Linguistics.

\bibitem[{Panayotov et~al.(2015)Panayotov, Chen, Povey, and
  Khudanpur}]{panayotov2015librispeech}
Vassil Panayotov, Guoguo Chen, Daniel Povey, and Sanjeev Khudanpur. 2015.
\newblock Librispeech: an {ASR} corpus based on public domain audio books.
\newblock In \emph{2015 IEEE international conference on acoustics, speech and
  signal processing (ICASSP)}, pages 5206--5210.

\bibitem[{Paszke et~al.(2019)Paszke, Gross, Massa, Lerer, Bradbury, Chanan,
  Killeen, Lin, Gimelshein, Antiga, Desmaison, Kopf, Yang, DeVito, Raison,
  Tejani, Chilamkurthy, Steiner, Fang, Bai, and
  Chintala}]{paszke-etal-2019-pytorch}
Adam Paszke, Sam Gross, Francisco Massa, Adam Lerer, James Bradbury, Gregory
  Chanan, Trevor Killeen, Zeming Lin, Natalia Gimelshein, Luca Antiga, Alban
  Desmaison, Andreas Kopf, Edward Yang, Zachary DeVito, Martin Raison, Alykhan
  Tejani, Sasank Chilamkurthy, Benoit Steiner, Lu~Fang, Junjie Bai, and Soumith
  Chintala. 2019.
\newblock \href
  {https://proceedings.neurips.cc/paper\_files/paper/2019/hash/bdbca288fee7f92f2bfa9f7012727740-Abstract.html}
  {{{PyTorch}}: {A}n imperative style, high-performance deep learning library}.
\newblock In \emph{Advances in {{Neural Information Processing Systems}}},
  volume~32.

\bibitem[{Prince and Smolensky(1997)}]{prince-1997-optimality}
Alan Prince and Paul Smolensky. 1997.
\newblock Optimality: From neural networks to universal grammar.
\newblock \emph{Science}, 275(5306):1604--1610.

\bibitem[{Radford et~al.(2019)Radford, Wu, Child, Luan, Amodei, Sutskever
  et~al.}]{radford-2019-gpt2}
Alec Radford, Jeffrey Wu, Rewon Child, David Luan, Dario Amodei, Ilya
  Sutskever, et~al. 2019.
\newblock Language models are unsupervised multitask learners.
\newblock \emph{OpenAI blog}, 1(8):9.

\bibitem[{Samuel et~al.(2023)Samuel, Kutuzov, {\O}vrelid, and
  Velldal}]{samuel-etal-2023-trained}
David Samuel, Andrey Kutuzov, Lilja {\O}vrelid, and Erik Velldal. 2023.
\newblock \href {https://doi.org/10.18653/v1/2023.findings-eacl.146} {Trained
  on 100 million words and still in shape: {BERT} meets {B}ritish {N}ational
  {C}orpus}.
\newblock In \emph{Findings of the Association for Computational Linguistics:
  EACL 2023}, pages 1954--1974, Dubrovnik, Croatia. Association for
  Computational Linguistics.

\bibitem[{Schatz et~al.(2021)Schatz, Feldman, Goldwater, Cao, and
  Dupoux}]{schatz2021early}
Thomas Schatz, Naomi~H Feldman, Sharon Goldwater, Xuan-Nga Cao, and Emmanuel
  Dupoux. 2021.
\newblock Early phonetic learning without phonetic categories: Insights from
  large-scale simulations on realistic input.
\newblock \emph{Proceedings of the National Academy of Sciences},
  118(7):e2001844118.

\bibitem[{Seidenberg and McClelland(1989)}]{seidenberg-1989-word-recognition}
Mark~S. Seidenberg and James~L. McClelland. 1989.
\newblock \href {https://doi.org/10.1037/0033-295X.96.4.523} {A distributed,
  developmental model of word recognition and naming}.
\newblock \emph{Psychological Review}, 96:523--568.

\bibitem[{Sennrich et~al.(2016)Sennrich, Haddow, and
  Birch}]{sennrich-etal-2016-bpe}
Rico Sennrich, Barry Haddow, and Alexandra Birch. 2016.
\newblock \href {https://doi.org/10.18653/v1/P16-1162} {Neural machine
  translation of rare words with subword units}.
\newblock In \emph{Proceedings of the 54th Annual Meeting of the Association
  for Computational Linguistics (Volume 1: Long Papers)}, pages 1715--1725,
  Berlin, Germany. Association for Computational Linguistics.

\bibitem[{Sundararaman et~al.(2021)Sundararaman, Kumar, and
  Vepa}]{sundararaman-2021-phonemebert}
Mukuntha~Narayanan Sundararaman, Ayush Kumar, and Jithendra Vepa. 2021.
\newblock \href {https://doi.org/10.21437/Interspeech.2021-1582} {{PhonemeBERT:
  Joint Language Modelling of Phoneme Sequence and ASR Transcript}}.
\newblock In \emph{Proc. Interspeech 2021}, pages 3236--3240.

\bibitem[{Suvarna et~al.(2024)Suvarna, Khandelwal, and
  Peng}]{suvarna-etal-2024-phonologybench}
Ashima Suvarna, Harshita Khandelwal, and Nanyun Peng. 2024.
\newblock \href {https://aclanthology.org/2024.knowllm-1.1}
  {{P}honology{B}ench: Evaluating phonological skills of large language
  models}.
\newblock In \emph{Proceedings of the 1st Workshop on Towards Knowledgeable
  Language Models (KnowLLM 2024)}, pages 1--14, Bangkok, Thailand. Association
  for Computational Linguistics.

\bibitem[{Suzgun et~al.(2023)Suzgun, Scales, Sch{\"a}rli, Gehrmann, Tay, Chung,
  Chowdhery, Le, Chi, Zhou, and Wei}]{suzgun-2023-Big-Bench}
Mirac Suzgun, Nathan Scales, Nathanael Sch{\"a}rli, Sebastian Gehrmann, Yi~Tay,
  Hyung~Won Chung, Aakanksha Chowdhery, Quoc Le, Ed~Chi, Denny Zhou, and Jason
  Wei. 2023.
\newblock \href {https://doi.org/10.18653/v1/2023.findings-acl.824}
  {Challenging {BIG}-bench tasks and whether chain-of-thought can solve them}.
\newblock In \emph{Findings of the Association for Computational Linguistics:
  ACL 2023}, pages 13003--13051, Toronto, Canada. Association for Computational
  Linguistics.

\bibitem[{Timiryasov and Tastet(2023)}]{timiryasov-tastet-2023-baby}
Inar Timiryasov and Jean-Loup Tastet. 2023.
\newblock \href {https://doi.org/10.18653/v1/2023.conll-babylm.24} {Baby
  {L}lama: knowledge distillation from an ensemble of teachers trained on a
  small dataset with no performance penalty}.
\newblock In \emph{Proceedings of the BabyLM Challenge at the 27th Conference
  on Computational Natural Language Learning}, pages 279--289, Singapore.
  Association for Computational Linguistics.

\bibitem[{{\"U}st{\"u}n et~al.(2018){\"U}st{\"u}n, Kurfal{\i}, and
  Can}]{ustun-etal-2018-characters}
Ahmet {\"U}st{\"u}n, Murathan Kurfal{\i}, and Burcu Can. 2018.
\newblock \href {https://doi.org/10.18653/v1/W18-3019} {Characters or
  morphemes: How to represent words?}
\newblock In \emph{Proceedings of the Third Workshop on Representation Learning
  for {NLP}}, pages 144--153, Melbourne, Australia. Association for
  Computational Linguistics.

\bibitem[{Wang et~al.(2019)Wang, Pruksachatkun, Nangia, Singh, Michael, Hill,
  Levy, and Bowman}]{wang-etal-2019-superglue}
Alex Wang, Yada Pruksachatkun, Nikita Nangia, Amanpreet Singh, Julian Michael,
  Felix Hill, Omer Levy, and Samuel Bowman. 2019.
\newblock \href
  {https://proceedings.neurips.cc/paper_files/paper/2019/file/4496bf24afe7fab6f046bf4923da8de6-Paper.pdf}
  {{SuperGLUE}: A stickier benchmark for general-purpose language understanding
  systems}.
\newblock In \emph{Advances in Neural Information Processing Systems},
  volume~32.

\bibitem[{Wang et~al.(2018)Wang, Singh, Michael, Hill, Levy, and
  Bowman}]{wang-etal-2018-glue}
Alex Wang, Amanpreet Singh, Julian Michael, Felix Hill, Omer Levy, and Samuel
  Bowman. 2018.
\newblock \href {https://doi.org/10.18653/v1/W18-5446} {{GLUE}: A multi-task
  benchmark and analysis platform for natural language understanding}.
\newblock In \emph{Proceedings of the 2018 {EMNLP} Workshop {B}lackbox{NLP}:
  Analyzing and Interpreting Neural Networks for {NLP}}, pages 353--355,
  Brussels, Belgium. Association for Computational Linguistics.

\bibitem[{Warstadt and Bowman(2022)}]{warstadt-2022-artificial}
Alex Warstadt and Samuel~R Bowman. 2022.
\newblock What artificial neural networks can tell us about human language
  acquisition.
\newblock In \emph{Algebraic structures in natural language}, pages 17--60. CRC
  Press.

\bibitem[{Warstadt et~al.(2023)Warstadt, Mueller, Choshen, Wilcox, Zhuang,
  Ciro, Mosquera, Paranjabe, Williams, Linzen, and
  Cotterell}]{warstadt-2023-babylm-findings}
Alex Warstadt, Aaron Mueller, Leshem Choshen, Ethan Wilcox, Chengxu Zhuang,
  Juan Ciro, Rafael Mosquera, Bhargavi Paranjabe, Adina Williams, Tal Linzen,
  and Ryan Cotterell. 2023.
\newblock \href {https://doi.org/10.18653/v1/2023.conll-babylm.1} {Findings of
  the {B}aby{LM} challenge: Sample-efficient pretraining on developmentally
  plausible corpora}.
\newblock In \emph{Proceedings of the BabyLM Challenge at the 27th Conference
  on Computational Natural Language Learning}, pages 1--34, Singapore.
  Association for Computational Linguistics.

\bibitem[{Warstadt et~al.(2020)Warstadt, Parrish, Liu, Mohananey, Peng, Wang,
  and Bowman}]{warstadt-etal-2020-blimp-benchmark}
Alex Warstadt, Alicia Parrish, Haokun Liu, Anhad Mohananey, Wei Peng, Sheng-Fu
  Wang, and Samuel~R. Bowman. 2020.
\newblock \href {https://doi.org/10.1162/tacl_a_00321} {{BL}i{MP}: The
  benchmark of linguistic minimal pairs for {E}nglish}.
\newblock \emph{Transactions of the Association for Computational Linguistics},
  8:377--392.

\bibitem[{Wolf et~al.(2020)Wolf, Debut, Sanh, Chaumond, Delangue, Moi, Cistac,
  Rault, Louf, Funtowicz, Davison, Shleifer, {von Platen}, Ma, Jernite, Plu,
  Xu, Le~Scao, Gugger, Drame, Lhoest, and Rush}]{wolf-etal-2020-transformers}
Thomas Wolf, Lysandre Debut, Victor Sanh, Julien Chaumond, Clement Delangue,
  Anthony Moi, Pierric Cistac, Tim Rault, Remi Louf, Morgan Funtowicz, Joe
  Davison, Sam Shleifer, Patrick {von Platen}, Clara Ma, Yacine Jernite, Julien
  Plu, Canwen Xu, Teven Le~Scao, Sylvain Gugger, Mariama Drame, Quentin Lhoest,
  and Alexander Rush. 2020.
\newblock \href {https://doi.org/10.18653/v1/2020.emnlp-demos.6} {Transformers:
  {S}tate-of-the-art natural language processing}.
\newblock In \emph{Proceedings of the 2020 {{Conference}} on {{Empirical
  Methods}} in {{Natural Language Processing}}: {{System Demonstrations}}},
  pages 38--45, {Online}. {Association for Computational Linguistics}.

\bibitem[{Xue et~al.(2022)Xue, Barua, Constant, Al-Rfou, Narang, Kale, Roberts,
  and Raffel}]{xue-etal-2022-byt5}
Linting Xue, Aditya Barua, Noah Constant, Rami Al-Rfou, Sharan Narang, Mihir
  Kale, Adam Roberts, and Colin Raffel. 2022.
\newblock Byt5: Towards a token-free future with pre-trained byte-to-byte
  models.
\newblock \emph{Transactions of the Association for Computational Linguistics},
  10:291--306.

\bibitem[{Zellers et~al.(2019)Zellers, Holtzman, Bisk, Farhadi, and
  Choi}]{zellers-etal-2019-hellaswag}
Rowan Zellers, Ari Holtzman, Yonatan Bisk, Ali Farhadi, and Yejin Choi. 2019.
\newblock \href {https://doi.org/10.18653/v1/P19-1472} {{H}ella{S}wag: Can a
  machine really finish your sentence?}
\newblock In \emph{Proceedings of the 57th Annual Meeting of the Association
  for Computational Linguistics}, pages 4791--4800, Florence, Italy.
  Association for Computational Linguistics.

\bibitem[{Zhu et~al.(2024)Zhu, Yang, Samir, and Islam}]{zhu-etal-2024-taste}
Jian Zhu, Changbing Yang, Farhan Samir, and Jahurul Islam. 2024.
\newblock \href {https://doi.org/10.18653/v1/2024.naacl-long.43} {The taste of
  {IPA}: Towards open-vocabulary keyword spotting and forced alignment in any
  language}.
\newblock In \emph{Proceedings of the 2024 Conference of the North American
  Chapter of the Association for Computational Linguistics: Human Language
  Technologies (Volume 1: Long Papers)}, pages 750--772, Mexico City, Mexico.
  Association for Computational Linguistics.

\end{thebibliography}

\newpage

\appendix

\section{Implementation Details}\label{app:implementation_details}

We implement all experiments using the \texttt{PyTorch} framework \citep{paszke-etal-2019-pytorch} and the \texttt{Transformers} library \citep{wolf-etal-2020-transformers}.

\subsection{Hardware Details}

We use a server with one NVIDIA A100 80GB PCIe GPU, 32 CPUs, and 32 GB of RAM for all experiments. Below, we report a subset of the output of the \emph{lscpu} command:

\begin{tcolorbox}[left=5pt,right=5pt,top=5pt,bottom=5pt]
\small
\begin{verbatim}
Architecture:        x86_64
CPU op-mode(s):      32-bit, 64-bit
Address sizes:       46 bits physical, 
                     48 bits virtual
Byte Order:          Little Endian
CPU(s):              32
On-line CPU(s) list: 0-31
Vendor ID:           GenuineIntel
Model name:          Intel(R) Xeon(R)
                     Silver 4210R CPU
                     @ 2.40GHz
CPU family:          6
Model:               85
Thread(s) per core:  1
Core(s) per socket:  1
Socket(s):           8
Stepping:            7
BogoMIPS:            4800.11
\end{verbatim}
\end{tcolorbox}

\subsection{Model Parameters and Training Procedure}

\begin{table}[h!]
    \centering
    \small
    \begin{tabular}{lc}
    \toprule
         Parameter & Value\\
    \midrule
         Layers & 12 \\
         Heads & 12 \\
         Dropout & 0.1 \\
         Embedding Size & 768 \\
         Inner Size & 3072 \\
         Max Example Length & 128 \\
         Learning Rate & 0.001 \\
         Optimizer & AdamW \\
         Scheduler Type & Linear\\
         Max Steps & 400,000 \\
         Warm-up Steps & 90,000\\
         Per Device Batch Size & 32 \\
    \bottomrule
    \end{tabular}
    \caption{Hyperparameter settings for training the GPT-2 architecture. Vocabulary size varies according to the tokenizer used, but all other parameters are constant across experiments. Where values are not reported, they may be assumed to be default values.}
    \label{table:baseline_hyperparams}
\end{table}

We describe the model and training parameters in \cref{table:baseline_hyperparams}. The model parameters were chosen to match those of the Pythia-170M model from the Pythia suite \citep{biderman2023pythia}. The model has 85M non-embedding parameters and is also equivalent in size to GPT-Neo 125M and OPT-125M. The Pythia models use the GPTNeoX architecture which is slightly different to GPT-2. In initial experiments, we found that GPT-2 performed better on the benchmarks across all eight of our conditions. 

Data is prepared into batches by first tokenizing the entire dataset, combining all tokens into one long vector, and then splitting the vector into chunks of 128 tokens. Only the very last example is padded, if required. At each step during training, random chunks are selected and combined into batches. 

Checkpoints are taken every 50,000 steps during training. At each checkpoint, the perplexity is evaluated on the held-back evaluation set, and at the end of training the checkpoint with the lowest perplexity is returned as the best model. 

\section{Evaluation Details}

\subsection{Significance Tests}
\label{sec:significance}

\begin{table*}[t]
    \centering
    \small
    \begin{tabular}{l|cccc}
     & BLiMP&BLiMP Supplement & GLUE & BabySLM (Syntactic) \\
\hline
orthographic vs.\ phonemic & \textbf{0.0001} & 0.0780 & \textbf{0.0149} & 0.1884 \\
word boundaries vs.\ no word boundaries & \textbf{0.0000} & 0.1831 & 0.0813 & \textbf{0.0118} \\
character tokenization vs.\ subword tokenization & 0.5069 & 0.4832 & \textbf{0.0010} & 0.1500 \\
    \end{tabular}
    \caption{$p$-values from the paired student t-tests for each experiment. Significant results are given in \textbf{bold} using an alpha level of 0.05.}
    \label{tab:pvalues}
\end{table*}

It is difficult to determine whether the results for a given benchmark are significant given that we only train a single run for each of the eight conditions. Instead, we calculate the significance of a particular transformation by comparing the scores for each subtask of a benchmark. We average the scores achieved by the four models with a transformation applied and average the scores achieved by the four models without the transformation applied, giving us paired results for each subtask. We then use a paired student $t$-test to assess the significance of the transformation. We give the $p$-values for our significance tests in \cref{tab:pvalues}.

Note that there are 67 subtasks for BLiMP, 5 for BLiMP Supplement, 9 for GLUE and 9 for BabySLM (Syntactic). With only 5 pairs for BLiMP Supplement, 
the test is under-powered and low $p$-values are unlikely. There are no subtasks for BabySLM (Lexical) so significance cannot be computed in the same way.

\subsection{The Effect of End-of-Sentence Tokens}
\label{sec:endofsentence}

By default, our tokenizers add a special start-of-sentence token \texttt{UTT\_BOUNDARY} to all sentences. This corresponds to the \texttt{<s>} token often used by tokenizers to help transformers with sentence-level processing, and also represents utterance boundaries, which unlike word boundaries are a clear cue present in speech and often included in word segmentation studies \citep{feliciano-de-faria-2019-utterance-boundaries}. 

Since sentences are collated together during training, this means that these tokens also appear at the end of every sentence, implicitly acting as end-of-sentence tokens. As a result, the model may use them to represent sentence-level information (especially given that these models are auto-regressive). However, in most evaluation tasks, sentences are presented individually (with padding) and so by default the tokenizer does not add this token to the end of sentences. 

\begin{table*}[t]
    \centering
    \small
    \begin{tabular}{rcc}
        \toprule
        & Grammatical & Ungrammatical \\
        \midrule
       Original & Patrick revealed what a lot of men wore. & Patrick revealed that a lot of men wore. \\
       \midrule
       BPE Text Tokenizer & \makecell{\mybox{<s>}~~\mybox{\textvisiblespace patrick}~~\mybox{\textvisiblespace revealed}~ \mybox{\textvisiblespace what}\vspace{2pt}\\ ~\mybox{\textvisiblespace a}~~\mybox{\textvisiblespace lot}~~\mybox{\textvisiblespace of}~~\mybox{\textvisiblespace men}~~\mybox{\textvisiblespace wore}~~\mybox{\textvisiblespace .}} & \makecell{\mybox{<s>}~~\mybox{\textvisiblespace patrick}~ \mybox{\textvisiblespace revealed}~ \mybox{\textvisiblespace that}\vspace{2pt}\\ ~\mybox{\textvisiblespace a}~~\mybox{\textvisiblespace lot}~~\mybox{\textvisiblespace of}~~\mybox{\textvisiblespace men}~~\mybox{\textvisiblespace wore}~~\mybox{\textvisiblespace .}} \\
       \midrule
       BPE Phoneme Tokenizer & \makecell{ \mybox{<s>}~~\mybox{\textvisiblespace \textipa{p\ae t\*rIk}}~~\mybox{\textvisiblespace \textipa{\*rIvi:ld}}~~\mybox{\textvisiblespace \textipa{w2t}}\vspace{2pt}\\~~\mybox{\textvisiblespace \textipa{2}}~~\mybox{\textvisiblespace\textipa{lAt}} ~~\mybox{\textvisiblespace \textipa{2v}}~~\mybox{\textvisiblespace \textipa{mEn}}~~\mybox{\textvisiblespace \textipa{wO\*r}}} & \makecell{ \mybox{<s>}~~\mybox{\textvisiblespace \textipa{p\ae t\*rIk}}~~\mybox{\textvisiblespace \textipa{\*rIvi:ld}}~~\mybox{\textvisiblespace \textipa{T\ae t}}\vspace{2pt}\\~~\mybox{\textvisiblespace \textipa{2}}~~\mybox{\textvisiblespace\textipa{lAt}} ~~\mybox{\textvisiblespace \textipa{2v}}~~\mybox{\textvisiblespace \textipa{mEn}}~~\mybox{\textvisiblespace \textipa{wO\*r}}} \\
       \bottomrule
    \end{tabular}
    \caption{An example sentence pair from the \texttt{wh\_vs\_that\_with\_gap} subtask in BLiMP and the outputted tokens from our two tokenizers that use subwords but do not remove word boundaries. The `\textvisiblespace ' character denotes word boundaries and the `<s>' token represents our \texttt{UTT\_BOUNDARY} token which acts as an utterance boundary and a start-of-sentence token.}
    \label{tab:blimpexample}
\end{table*}

This has consequences for zero-shot evaluation tasks where the grammaticality of the sentence depends on the sentence being marked as complete, which is the case for several of the BLiMP subtasks. For instance, one subtask evaluates a model's understanding of filler-gap dependencies by presenting grammatical ``wh''-phrases with ``that''-phrases that are ungrammatical due to a missing dependency. An example is given in \cref{tab:blimpexample} along with the tokens produced by two of our tokenizers. Crucially, our phonemic transcriptions do not include punctuation (see \cref{sec:punctuation}) and for this task, without an end-of-sentence marker, the ``ungrammatical'' sentence is no longer ungrammatical, as it could just be incomplete.

This means that the subtask remained a valid test for our orthographic models (due to the inclusion of punctuation to mark the end of the sentence), but not the phonemic ones, since for the phonemic models both the ``grammatical'' and ``ungrammatical'' sentences could be considered grammatical. Since this task is not balanced, any preference for the word ``that'' over the ``wh''-words would lead to the model consistently choosing the ``that'' sentences and achieving results below chance (which is 0.5 for all BLiMP tasks).

In our initial experiments we found that the models trained on phonemes achieved scores between 0.06 and 0.14 for this task whereas the orthographic models achieved scores between 0.35 and 0.53. We then added the \texttt{UTT\_BOUNDARY} token to the end of every evaluation instance and found that the phonemic models could then achieve scores between 0.26 and 0.34 (with little change for the orthographic models). These results also held for several other BLiMP tasks with similar constructions. 

We thus decided to ensure that the token was added to the end of every evaluation instance for all benchmarks reported in this paper for two reasons. First, it acts as a necessary end-of-sentence marker to ensure certain tests remain valid for the phonemic models, and second, because the token may encode useful sentence-level information for all models (particularly for GLUE tasks, as only the encoding of the final token is used for predictions).

We present the effect of this decision in \cref{fig:endofsentencetoken} which reports the overall BLiMP scores for our eight conditions with and without the inclusion of the \texttt{UTT\_BOUNDARY} token at the end of each evaluation sentence. There is a very large increase for all four phonemic models with little change for the orthographic models, confirming that this was a crucial adjustment.

\subsection{BabySLM Comparison}
\label{sec:babyslmcomparison}

In \cref{table:results} we report the BabySLM scores achieved by our models and in \cref{sec:babyslm} we mention that these are the highest scores achieved on this benchmark to date. It is worth noting that this is only in comparison to the baseline scores released with the BabySLM benchmark \citep{lavechin}, as at the time of writing no other scores have been published for this benchmark, given how recently it was introduced.

In their study, \citet{lavechin} achieved their highest syntactic score of 70.4 using BabyBERTa \citep{huebner-etal-2021-babyberta} trained on only 5 million words from CHILDES \citep{macwhinney1985child}. All of our models beat this score, with the highest achieving 94.9. BabyBERTa also uses a BPE tokenizer whereas we found that a character-based tokenizer consistently gave better performance (see \cref{sec:babyslm}). There is also an architectural difference, BabyBERTa is an autoencoder trained using masked language modeling, whereas our model is autoregressive, using next-token prediction. The LTG-BERT baseline, which is a similarly sized model also trained on 100 million words, only achieves a score of 75.8. The Baby Llama baseline, by comparison, achieves 94.0. It is possible that the autoregressive architecture is much more suited to the syntactic task than the autoencoder architecture of BERT. 

When it comes to the lexical test, the highest score achieved by \citet{lavechin} was 75.4 using a 3-layer LSTM trained on 1.2 million words from the Providence corpus \citep{borschinger-etal-2013-joint} which they converted to a stream of phonemes with no word boundaries using a similar tool to ours. Our highest-scoring model was also trained with character-based tokenization of phonemes, but did include word boundaries, achieving a score of 89.6. Our model without word boundaries got the second-highest score with 87.8.

In both cases, our model is larger (12 layers) and trained on much more data (100 million words) than the BabySLM baselines. Also, our pre-training dataset contains a wider variety of sentences than just the child-directed utterances in CHILDES. We are currently investigating the effect of model size and training size on the BabySLM scores. In initial experiments, we found that even a 6-layer model trained on only 7 million words from CHILDES was able to achieve a lexical score of 82, but this model also only achieved a syntactic score of 70. We hypothesize that lexical-level knowledge can be learned with less data and by smaller models when compared to learning syntactic knowledge, but this research is ongoing.



\section{Further Implications}
\label{sec:further}

\subsection{Comparing Human Acquisition to Language Model Learning}
\label{sec:acquisition}


The capacity of LMs to learn language from text alone has spurred interest in using such models for acquisition and psychology studies, such as comparing model learning trends to child learning behaviour \citep{evanson-etal-2023-language} and using model outputs to predict human reading times \citep{hollenstein-etal-2021-multilingual}.

To push this research further, recent efforts aim to make language modeling more cognitively plausible \citep{beinborn2024cognitive} by reducing the advantages that typical language models have over humans during the learning process \citep{warstadt-2022-artificial}. One approach is to limit and curate the dataset to that which a typical human may be exposed to, such as is done in the BabyLM challenge \citep{warstadt-2023-babylm-findings}. Another approach is to use an input representation that more closely mimics speech rather than written text \citep{dupoux-2018-cognitive}. Finally, we must consider whether the architectures themselves are suitable linguistic theories, given that they were developed for downstream tasks \citep{baroni-2022-proper}.

In this work we contribute to all three approaches by training a language model with streams of phonemes and assess whether the language model architecture used is advantaged or disadvantaged by these changes according to a wide variety of benchmarks. We hope that this leads to further work studying acquisition using phoneme streams as an input representation. However, while streams of phonemes may seem more cognitively plausible than written text, many studies go further than we do and seek to train directly on raw audio.

\subsection{Learning directly from audio}
\label{sec:audiomodels}

Our study focused on alternative input representations for text-based language models, but there is also a field of work dedicated to training models directly from raw audio.
In recent years, the Zero Resource Speech Challenge has helped pioneer the development of models that learn unsupervised from raw audio \citep{dunbar_self-supervised_2022}. Models such as STELA \citep{schatz2021early, lavechin2022can} use a two-stage approach, learning a discrete symbolic representation by clustering 10ms chunks of audio, then feeding these to a multi-layered LSTM language model.

These models are also used to study acquisition, regarding raw audio as an input representation that is more cognitively plausible than phonemes; a continuous signal full of noise and non-linguistic information that children must learn to filter. Whether adults even use phonemes as a core linguistic representation, and whether children learn phonemic categories before other stages of acquisition both continue to be a matter of debate \citep{kazanina2018phonemes, matusevych2023infant} and the symbolic representations learned by models such as STELA have a duration four times shorter than phonemes, challenging the assumption that phonemic categories are precursors to later stages of acquisition. 

The gap in linguistic performance between text-based models and audio-based models continues to be substantial. \citet{lavechin} developed BabySLM to compare text-based models to speech-based models and highlighted this gap, but further noted that even speech-based models may not always train on plausible input, many often using audiobooks as their training data \citep{kahn2020libri}. When training the STELA model on 1024 hours of ecological long-form child-centered audio compared to 1024 hours of audiobooks, \citet{lavechin} found that the model trained on long-form audio achieved chance-level syntactic and lexical capabilities, highlighting how far we are from producing architectures that can learn from the same signals as human children.

\end{document}